\documentclass{article}

\usepackage[nonatbib,preprint]{neurips}

\usepackage[numbers]{natbib}

\usepackage{makecell}
\usepackage{amsmath}
\usepackage[utf8]{inputenc}
\usepackage[T1]{fontenc}  
\usepackage{xcolor}
\usepackage{adjustbox}
\usepackage{enumitem}
\usepackage{tocloft}

\usepackage[
  colorlinks=true,
  linkcolor=gray!40!black, 
  citecolor=blue!40!black,  
  urlcolor=blue!40!black 
]{hyperref}

\usepackage{url}          
\usepackage{booktabs}   
\usepackage{amsfonts}     
\usepackage{nicefrac}   
\usepackage{pifont}
\usepackage{tablefootnote} 
\usepackage{graphicx}
\usepackage{xspace}
\usepackage{wrapfig}
\usepackage{amsmath}
\usepackage{multirow}
\usepackage{subcaption}

\usepackage[capitalize]{cleveref}
\crefname{section}{Sec.}{Secs.}
\Crefname{section}{Section}{Sections}
\crefname{table}{Tab.}{Tabs.}
\Crefname{table}{Table}{Tables}
\crefname{figure}{Fig.}{Figs.}
\Crefname{figure}{Figure}{Figures}
\crefname{equation}{Eq.}{Eqs.}
\Crefname{equation}{Equation}{Equations}
\usepackage{amsfonts,amssymb}

\usepackage[utf8]{inputenc}
\usepackage{amsmath}
\usepackage{algorithm}
\usepackage{algorithmic}

\usepackage{caption}
\usepackage{setspace}

\makeatletter
\DeclareRobustCommand\onedot{\futurelet\@let@token\@onedot}
\def\@onedot{\ifx\@let@token.\else.\null\fi\xspace}

\def\eg{\emph{e.g}\onedot} 
\def\ie{\emph{i.e}\onedot}

\def\Ours{Emu3.5\xspace}
\def\Oursimage{Emu3.5\xspace}
\def\Ursa{Discrete Diffusion Adaptation\xspace}
\def\Ursaabbr{DiDA\xspace}

\title{\Ours: Native  Multimodal Models are World Learners}

\author{
Emu3.5 Team \\[3.5pt]
BAAI 
}

\begin{document}

\maketitle

\vspace{-2.8em} 
\begin{center}
\url{https://emu.world}
\end{center}

\begin{abstract}

We introduce \Ours, a large-scale multimodal world model that natively predicts the next state across vision and language.
\Ours is pre-trained end-to-end with a unified next-token prediction objective on a corpus of vision-language interleaved data containing over 10 trillion tokens, primarily derived from sequential frames and transcripts of internet videos.
The model naturally accepts interleaved vision-language inputs and generates interleaved vision-language outputs.
\Ours is further post-trained with large-scale reinforcement learning to enhance multimodal reasoning and generation.
To improve inference efficiency, we propose \Ursa (\Ursaabbr), which converts token-by-token decoding into bidirectional parallel prediction, accelerating per-image inference by about $20\times$ without sacrificing performance.
\Ours exhibits strong native multimodal capabilities, including long-horizon vision-language generation, any-to-image (X2I) generation, and complex text-rich image generation.
It also exhibits generalizable world-modeling abilities, enabling spatiotemporally consistent world exploration and open-world embodied manipulation across diverse scenarios and tasks.
For comparison, \Ours achieves performance comparable to Gemini 2.5 Flash Image (Nano Banana) on image generation and editing tasks and demonstrates superior results on
a suite of interleaved generation tasks. 
We open-source \Ours at \url{https://github.com/baaivision/Emu3.5} to support community research.

 \vskip 0.5em
 \begin{figure}[htbp]
     \centering
     \includegraphics[width=1.\linewidth]{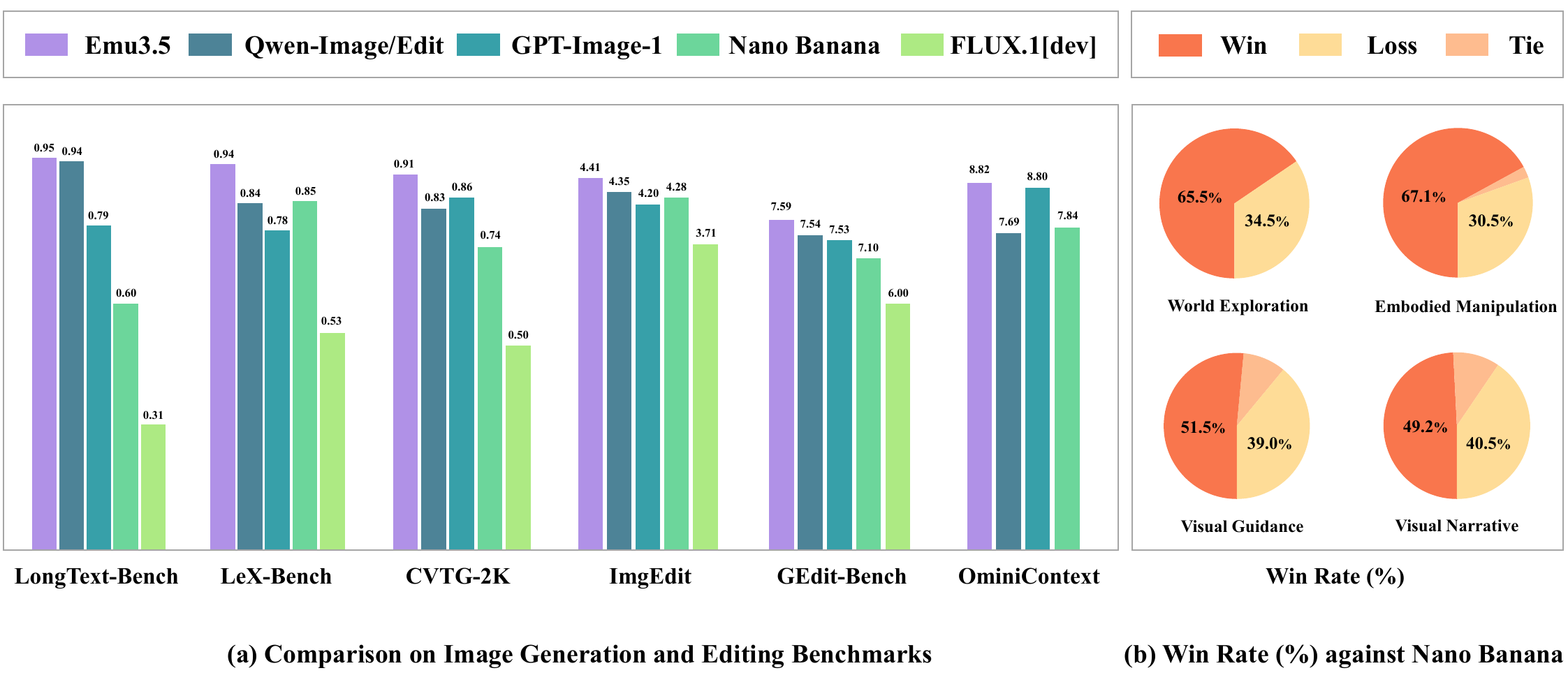}
      \caption{(a) Comparison with SOTA models on image generation and editing benchmarks. For editing, the specific models are Qwen-Image-Edit-2509~\citep{wu2025qwen} and FLUX.1 Kontext [dev]~\cite{labs2025flux}. (b) Automated preference evaluation (Win Rate[$\%$]) against Gemini 2.5 Flash Image (Nano Banana) \citep{gemini2p5} on interleaved generation tasks.}
     \label{fig:x2i_zft}
     \vspace{-1em}
 \end{figure}

\newgeometry{top=0.75cm,bottom=2cm}
\begin{figure}[htbp]
    \centering
    \makebox[\textwidth][c]{%
        \includegraphics[width=1.13\linewidth]{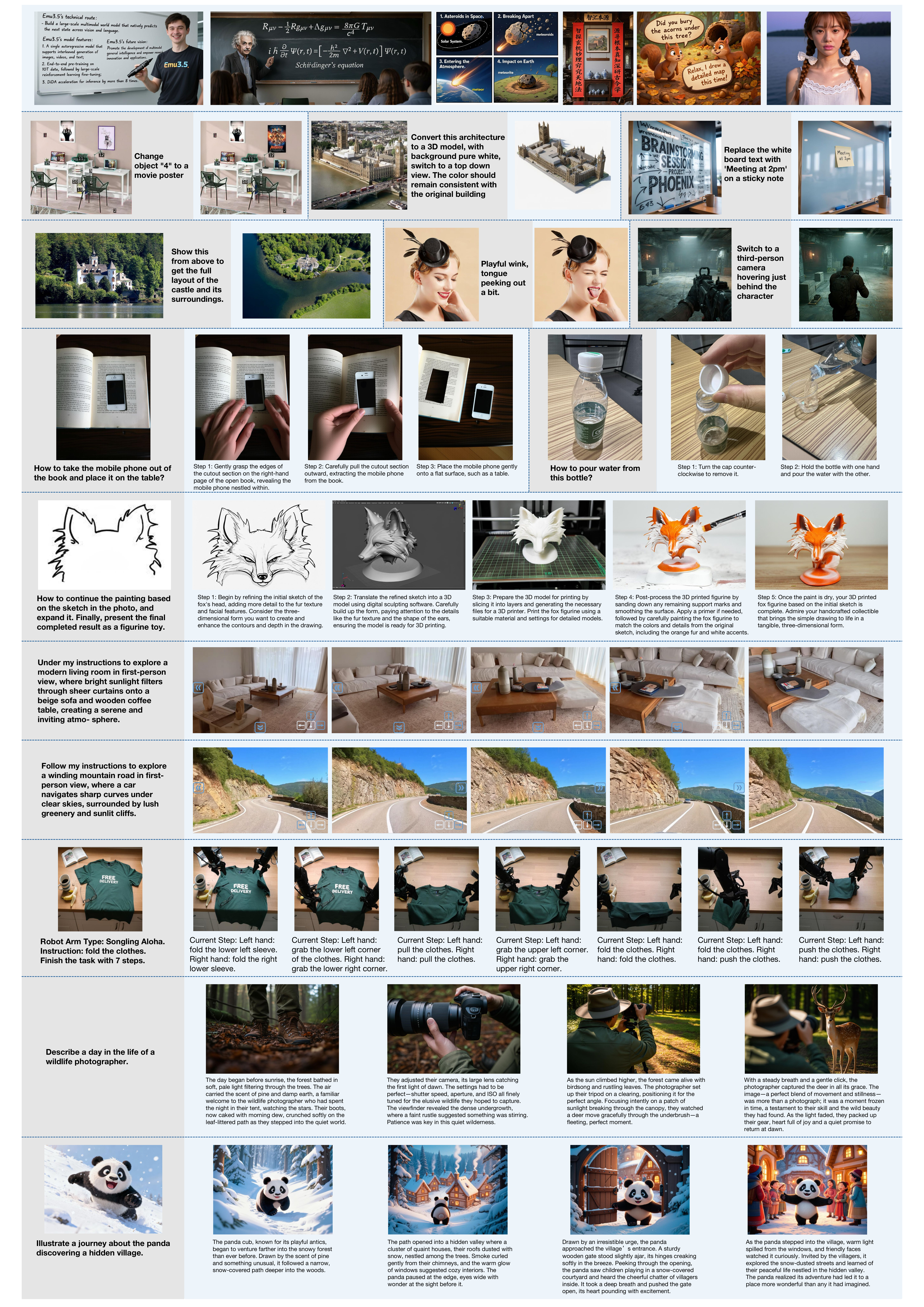}
    }
    \caption{
Native multimodal capabilities of \Ours. 
Gray: prompts; Blue: results.
}
    \label{fig:teaser}
\end{figure}
\restoregeometry

\end{abstract}

\newpage

\setlength{\cftbeforesecskip}{8pt}      
\setlength{\cftbeforesubsecskip}{6pt}    
\setlength{\cftbeforesubsubsecskip}{4pt} 
\begin{spacing}{0.9}
\tableofcontents
\end{spacing}

\newpage

\section{Introduction}

Language models trained on massive text corpora have achieved remarkable success in linguistic reasoning and generation~\cite{achiam2023gpt, anthropic2023claude35, gemini2p5, deepseekr1}, yet text alone provides only a limited view of the world. 
While language enables communication and generalization across people, vision is the primary modality through which humans perceive, interact with, and learn from the environment. 
Humans acquire knowledge not only from language, but also from spatially and temporally extended multimodal experiences, particularly long videos interleaved with language that encode rich context, causality, and temporal consistency.
Recent advances in short-clip video generation have demonstrated the ability to capture short-term dynamics, but learning from and reasoning over long-horizon vision-language sequences remains a central open challenge.

The previous Emu series~\cite{sun2023emu, emu2, wang2024emu3} demonstrated the feasibility of unifying multimodal tasks and modeling interleaved vision-language sequences through a simple generative objective, \eg, next-token prediction.
However, these efforts primarily focused on short-form or small-scale data,  leaving open fundamental questions about how to scale pre-training, post-training, and inference to handle long-horizon multimodal data.
In particular, it remains unclear how to effectively learn long videos interleaved with text, how to enable general-purpose multimodal interaction, and how to efficiently predict tens of thousands of visual tokens, which pose stringent demands on pre-training, post-training, and inference, respectively.

In this work, we address these challenges and build a world model that natively predicts the next state across interleaved vision and language. 
Specifically, we introduce \Ours, a large-scale multimodal world model trained to learn from and generalize over long-horizon multimodal data. 
\Ours is pre-trained end-to-end with a unified next-token prediction objective on a corpus of interleaved vision-language data containing over 10 trillion tokens, primarily sourced from sequential frames and transcripts of internet videos. 
For post-training, \Ours undergoes large-scale reinforcement learning guided by multimodal rewards for long-horizon generation. 
The model naturally processes interleaved inputs and generates interleaved outputs, enabling general-purpose multimodal reasoning. 
To improve the efficiency during inference, we propose Discrete Diffusion Adaptation (DiDA), which converts token-by-token decoding into bidirectional parallel prediction, accelerating per-image inference by approximately $20\times$ without sacrificing performance.

\Ours represents the first step toward large-scale native vision-language generation.
It demonstrates long-horizon multimodal generation and reasoning capabilities, producing interleaved sequences of visual frames and text that jointly capture temporal consistency and semantic coherence across chain of frames.
These capabilities enable a diverse range of tasks such as visual narrative and visual guidance, the former supporting coherent visual storytelling across open topics, including educational and imaginative narratives, and the latter enabling temporally consistent, step-by-step reasoning for illustrating complex procedures or tasks.
\Ours further exhibits generalizable world-modeling abilities encompassing world exploration and embodied manipulation, enabling controllable interaction, free-form navigation, and dynamic scene simulation across both real and imagined environments.
We carefully evaluate these new capabilities and demonstrate clear superiority of \Ours, a single 32B unified model, over the closed-source Gemini 2.5 Flash Image~\cite{google2025gemini25flashmodelcard}.

\Ours also serves as a state-of-the-art any-to-image (X2I) and text-to-image generation model, benefiting from its strong native multimodal capabilities.
 For any-to-image (X2I), \Ours enables open-world editing with precise control and free spatiotemporal manipulation.
 For image generation, it produces accurate, controllable, and natural text rendering.
The model supports multiple images as input and generating outputs up to 2K resolution.
 In comparison, Emu3.5 achieves performance comparable to Gemini 2.5 Flash Image on any-to-image (X2I) and surpasses it on text rendering.
Notably, it is also the first autoregressive model to rival closed-source diffusion models in both inference speed and generation quality.

We further make several insightful observations.
First, as pre-training compute scales up, the validation loss on out-of-distribution multimodal tasks continues to decrease, suggesting progressively stronger generalization beyond the training domains.
Second, unified post-training such as reinforcement learning, establishes a shared multimodal interface through which different tasks can mutually benefit and transfer; for example, the high fidelity of text-to-image generation and the editing ability in any-to-image tasks naturally transfer to visual narrative and visual guidance tasks.
Third, we find that the next-token prediction model can be efficiently transformed into a bidirectional predictor, achieving substantial acceleration without sacrificing performance.
Together, these observations highlight the scalability, versatility, and flexibility of the native multimodal paradigm.

We open-source \Ours to support community research and development.
The model natively enables an interactive interface for step-by-step vision-language interaction and serves as a foundation for developing new multimodal capabilities. We hope \Ours will pave the way toward advancing world models and improving multimodal intelligence.

\begin{figure}[t]
    \centering
    \includegraphics[width=1.0\linewidth]{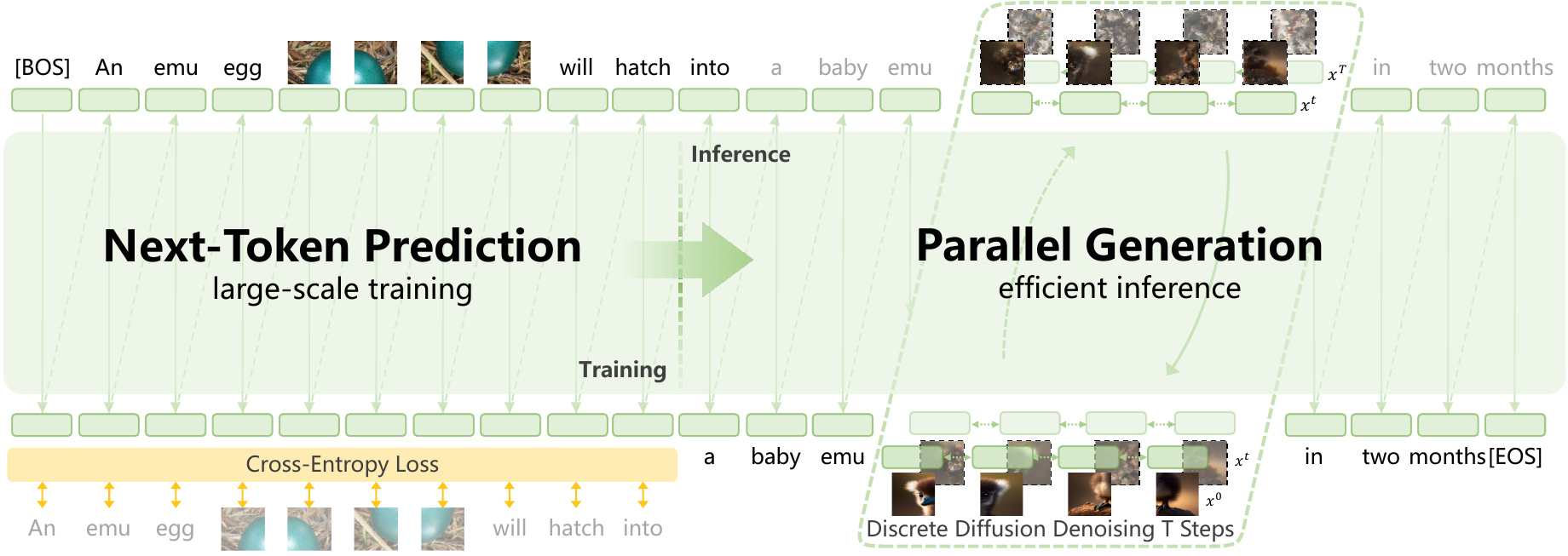}
     \caption{Overview of the \Ours architecture. The model is trained end-to-end at scale with a unified next-token prediction objective. During inference, single-token prediction is accelerated via discrete diffusion adaptation, enabling bidirectional parallel generation per image.}
    \label{fig:overall_arch}
    \vspace{-1em}
\end{figure}

\section{\Ours}

\subsection{Overall}

Figure~\ref{fig:overall_arch} illustrates the overall architecture of \Ours during both large-scale training and efficient inference. 
During training, the model performs unified next-token prediction (NTP) and follows a standard decoder-only transformer architecture for large-scale multimodal pre-training, supervised fine-tuning, and reinforcement learning. 
During inference, the proposed \Ursaabbr approach enables efficient hybrid generation, \ie, sequential textual generation and parallel visual generation, achieving $20\times$ acceleration per image without sacrificing quality.

\begin{figure}[t]
    \centering
    \includegraphics[width=0.999\linewidth]{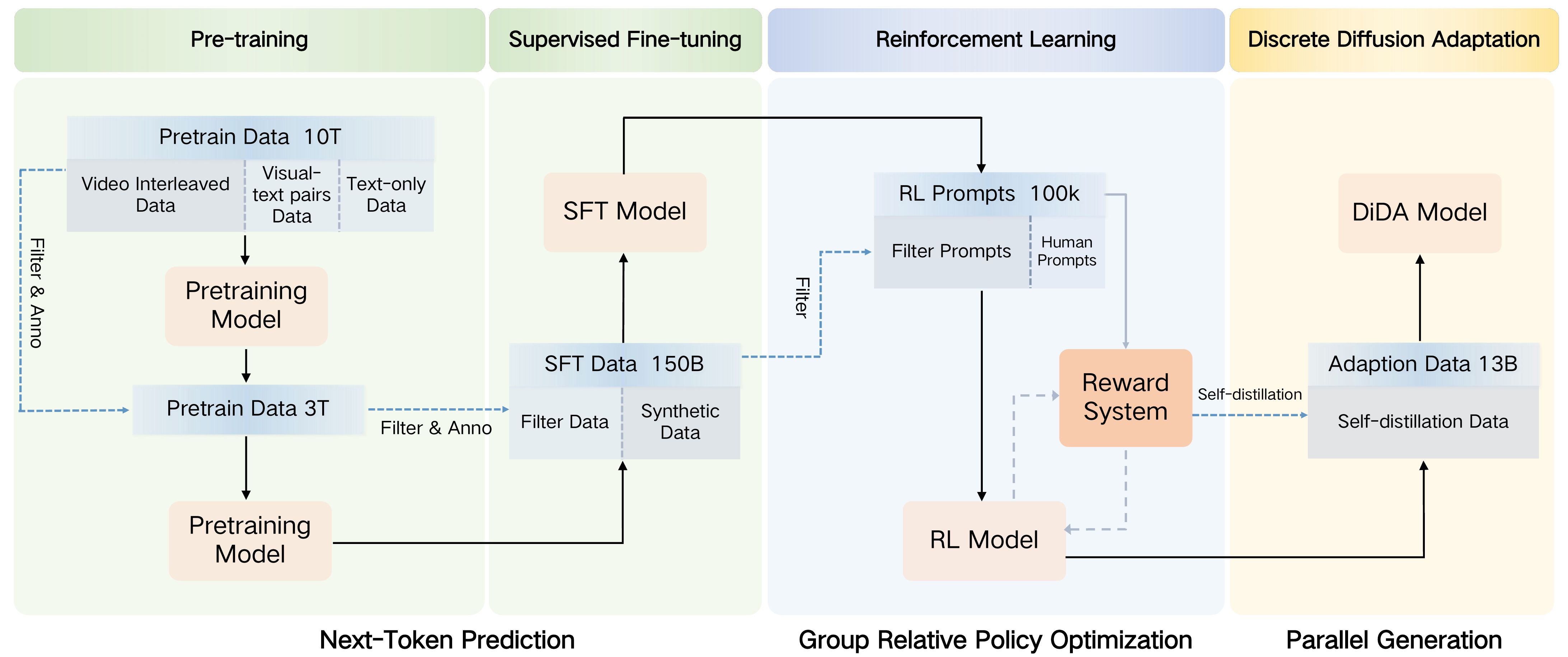}
     \caption{Overall training pipeline of \Ours.}
    \label{fig:overall_pipeline}
    \vspace{-1em}
\end{figure}

The complete training pipeline is illustrated in Figure~\ref{fig:overall_pipeline}. \Ours is first pre-trained end-to-end in two stages on approximately 13 trillion tokens, primarily derived from sequential frames and transcripts of internet videos. The second stage further improves visual resolution diversity, data quality, and annotation richness, providing more precise multimodal supervision. This two-stage setup enables the model to naturally process interleaved vision-language inputs and generate interleaved outputs within a unified generative framework. 
Subsequently, \Ours undergoes supervised fine-tuning (SFT) with 150 billion samples to establish a unified multimodal generation interface, followed by large-scale reinforcement learning to further enhance multimodal reasoning and generation capabilities. 
The model is then rapidly adapted for high-efficiency inference with \Ursaabbr, using only a few billions tokens from SFT and self-distillation data.

\subsection{Unified Architecture}

\Ours follows the standard transformer-based architecture commonly adopted in recent large language models such as Qwen3~\cite{touvron2023llama2}, while incorporating several design modifications to balance scalability and multimodal adaptability. The model consists of 64 transformer layers, each with a hidden size of 5,120 and an intermediate size of 25,600. The attention mechanism employs 64 heads with 8 dedicated key-value heads, adopting Grouped Query Attention (GQA)~\cite{ainslie2023gqa} to improve efficiency. RMSNorm~\cite{zhang2019root} with pre-normalization is used to stabilize training. 
We introduce QK-Norm~\cite{dehghani2023scaling} to the query and key projections to enhance attention stability. SwiGLU~\cite{shazeer2020glu} is used as the activation function, and rotary positional embeddings (RoPE)~\cite{su2024roformer} are employed.
Overall, the model contains 34.1 billion(B) parameters, including 31.2 B in the transformer layers and 2.9 B in the embedding layers. The total vocabulary size is 282,926, consisting of 151,854 text tokens and 131,072 vision tokens. The text vocabulary directly reuses QwenTokenizer\footnote{\url{https://huggingface.co/Qwen/Qwen-7B/blob/main/tokenization\_qwen.py}}, ensuring robust multilingual text coverage. The visual vocabulary is learned from diverse images and will be detailed in Section~\ref{sec:vtokenizer}. The model supports a context length of up to 32,768 tokens and applies a dropout rate of 0.1 to stabilize training. Detailed model configurations are summarized in Table~\ref{tab:model_configurations}.

\begin{table}[htbp]
\centering
\small
\begin{tabular}{cccccccc}
\toprule
 \textbf{Parameters(B)} & \textbf{Layers}  & \textbf{Hidden Size} & \textbf{Intermediate Size} & \textbf{Heads (Q / KV)} & \textbf{Vocabulary Size}  & \textbf{Context Length} \\
\midrule
 34.1 & 64 & 5120 & 25600 & 64/8 & 282926  & 32768 \\
\bottomrule
\end{tabular}
\vspace{0.5em}
\caption{Model configurations of \Ours.}
\label{tab:model_configurations}
\end{table}

\subsection{Tokenizer}
\label{sec:vtokenizer}
We primarily adopt the IBQ~\cite{VQ:IBQ} framework for visual tokenization with a downsampling factor of $f=16$. Each discrete token in the codebook has a dimension of $D=256$. To further increase the tokenizer’s capacity, we expand the codebook size to 131,072, and the model size is also increased to 455 million parameters through width scaling, enhancing its ability to represent complex image structures. Inspired by REPA~\cite{VQ:REPA}, we also integrate feature distillation from SigLIP~\cite{SigLIP} into the intermediate outputs of the tokenizer decoder during training, improving representation learning and enriching the semantic information of the discrete image tokens.

\vspace{-0.015\linewidth}
\paragraph{Image decoder.\!\!\!}
Our vanilla tokenizer achieves superior reconstruction quality while using only one-fourth of the tokens required by Emu3 to represent a same image.
To further enhance visual decoding, we introduce diffusion-based decoders as an optional alternative to the vanilla image decoder. 
The diffusion-based image decoder takes the same quantized tokens as input but generates images at twice the resolution of the vanilla decoder. It improves local details and fine-grained details, particularly in text regions and facial reconstruction.
Moreover, following~\cite{chadebec2025flash}, we perform the LoRA-based distillation method to accelerate the decoding by about 10$\times$, \ie, from $50$ denoised steps to $4$, without sacrificing performance.

\vspace{-0.015\linewidth}
\paragraph{Video decoder.\!\!\!} We extend \Ours to generate continuous videos with a diffusion-based video decoder conditioned on the generated keyframe tokens.
Our video decoder is built upon the mainstream DiT~\citep{Peebles2022DiT} architecture. 
We utilize quantized embeddings from the VQ quantizer to provide fine-grained visual details, while optional inter-frame textual information is used to supply high-level semantic guidance.
We further introduce an additional 4-channel mask to indicate which frames' tokens are provided, enabling the model to support an arbitrary number of intermediate frames. During training, we randomly replace the first keyframe latent with clean image tokens to bridge long-term temporal dependencies and enhance its generalization across diverse keyframe conditions.

\section{Pre-training}

\subsection{Training Data}

The pre-training data of \Ours\ comprises over 13 trillion multimodal tokens, representing an advancement over Emu3~\cite{wang2024emu3} in terms of scale, diversity and quality. 
Our pre-training dataset integrates four major components: (1) interleaved vision-language data, (2) vision-text pairs, (3) any-to-image data, and (4) text-only data. 

\subsubsection{Video Interleaved Data}

\begin{figure}[t]
    \centering
    \includegraphics[width=0.99\linewidth]{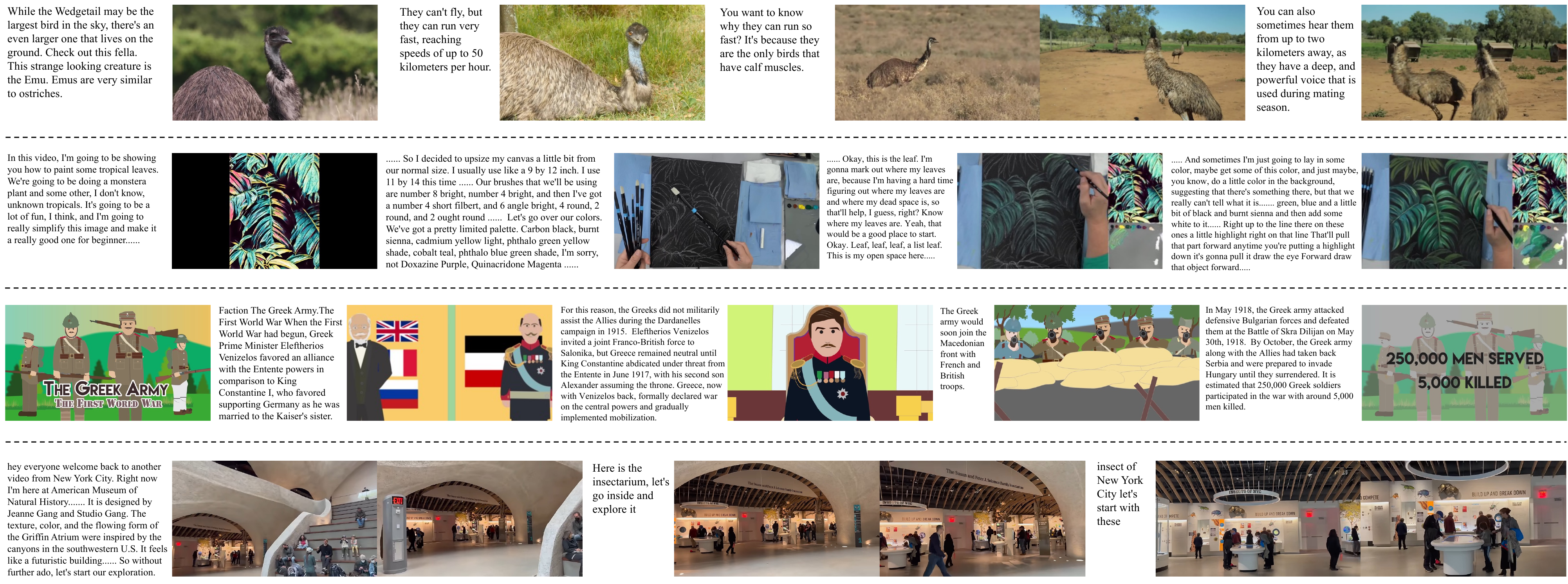}
     \caption{Video interleaved data samples from \Ours's pre-training dataset.}
     \label{subfig:data_cases} 
\end{figure}

Unlike conventional approaches~\cite{qwen2p5vl,llava-onevision,llava-1.5,QWEN2VL,wu2025qwen} that primarily rely on paired data composed of short, independent samples, our corpus is constructed to capture long-horizon, interleaved multimodal context. Specifically, this subset is derived from sequential video frames and temporally aligned audio transcripts of large-scale internet videos, which inherently preserve spatiotemporal continuity, cross-modal alignment, and contextual coherence. 
Figure~\ref{subfig:data_cases} presents several examples of video-interleaved data. This type of long-horizon multimodal sequence provides substantially richer context than isolated pairs and facilitates the model’s learning of extended-horizon generation, reasoning, and world modeling over extended temporal spans.

\begin{figure}[t]
    \centering
    \includegraphics[width=0.99\linewidth]{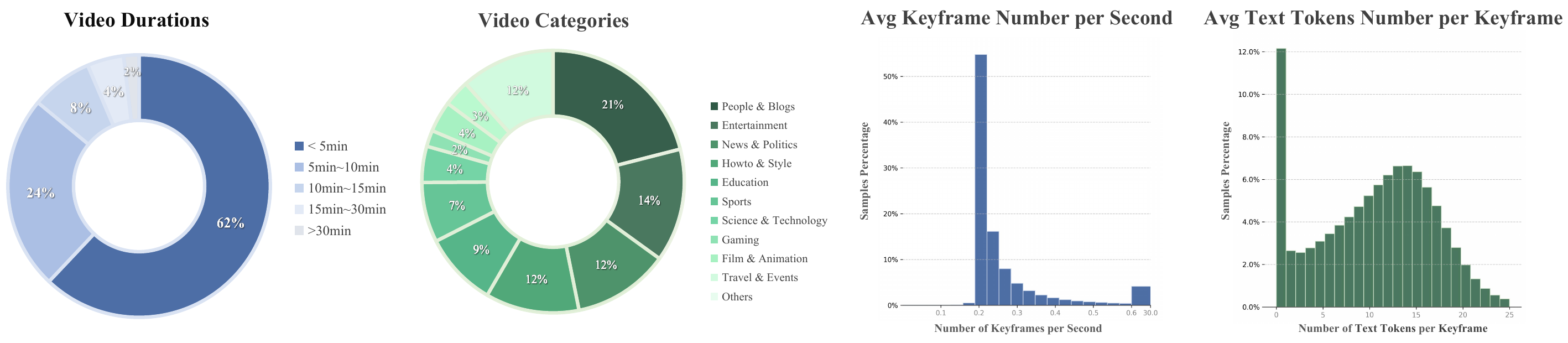}
     \caption{Data statistics of video interleaved data.}
    \label{subfig:data_static}
\end{figure}

\textbf{Data Collection}. Our interleaved vision-language data is sourced from diverse videos, including open-source datasets, publicly available online videos, and videos accessed through partnerships with third parties.
In total, the corpus comprises approximately 63 million videos with an average duration of 6.5 minutes, amounting to roughly 790 years of continuous footage. 
The collected corpus spans numerous domains such as education, science and technology, How-to, entertainment, sports, gaming, travel, and animation, thereby capturing a comprehensive spectrum of both real-world and imaginative scenarios. 
Figure~\ref{subfig:data_static} presents the statistical distribution of video duration and video categories of the collected interleaved video data.
This design capitalizes on the inherently scalable nature of internet video content, similar to web-scale textual corpora, allowing the dataset to be continuously expanded across domains, tasks, and scenarios.

\textbf{Data Preprocessing.}
The raw video data, which contains both visual frames and audio tracks, requires separate yet coordinated preprocessing steps. 
We preprocess video frames by first employing PySceneDetect\footnote{\url{https://github.com/Breakthrough/PySceneDetect}} to segment each video into coherent scenes. For each detected scene, a single middle frame is selected if its duration is shorter than $t$ seconds; otherwise, frames are sampled every $t$ seconds, with timestamps recorded. Empirical analysis shows that this strategy better preserves the essential visual content of the video while effectively removing redundant frames, outperforming uniform sampling. We analyzed the distribution of keyframes extracted per second, as shown in Figure~\ref{subfig:data_static}. Overall, the extraction intervals are relatively consistent, with an average of 0.27 keyframes per second.

For the audio track, we adopt the Whisper-large-v2 model~\cite{whisper} to perform automatic speech recognition (ASR), accelerated by the Faster-Whisper\footnote{\url{https://github.com/SYSTRAN/faster-whisper}} implementation. The resulting transcripts and word-level timestamps are further refined through post-processing using the spaCy\footnote{\url{https://github.com/explosion/spaCy}}, which segments text based on temporal pauses and syntactic rules to produce grammatically coherent and temporally aligned transcriptions. We counted the average number of ASR text tokens corresponding to each keyframe per video, which is shown in Figure~\ref{subfig:data_static}. We computed the average number of ASR text tokens per keyframe for each video, as shown in Figure~\ref{subfig:data_static}. The distribution is relatively balanced; however, there is a certain amount of silent videos in the data, leading to a large number of cases with no text tokens, which will be balanced in subsequent data processing.

Finally, the integration of extracted keyframes with processed ASR transcripts yields naturally interleaved video-text sequences ordered by timestamps, providing a rich contextual structure for multimodal pre-training. Figure~\ref{subfig:data_cases} presents several examples of the processed video-text sequences.

\textbf{Data Filtering.} 
To guarantee the overall quality and consistency of the interleaved corpus, we design a two-stage filtering pipeline consisting of \textbf{basic filtering} and \textbf{advanced filtering}. The basic filtering stage is applied during the first-phase pre-training, while the combination of both stages is employed in the second-phase pre-training, as detailed in later sections.

\begin{itemize}
    \item \textbf{Basic Filtering.}
    This stage performs coarse-level data cleaning and sample balancing. It includes:
        (1) Duration and resolution filtering: Videos with extremely short duration or low resolution are excluded to maintain stable visual quality.  
        (2) Talking-head filtering: We identify and exclude talking-head videos by combining a face detection model~\cite{deng2020retinaface} with Qwen-VL-based classification.  
        (3) Language and silence balancing: ASR transcripts are analyzed to detect multilingual content and silent videos, ensuring a balanced distribution across languages and reducing the proportion of silent videos.

    \item \textbf{Advanced Filtering.}  
    This stage refines the dataset through multimodal quality evaluation and redundancy reduction, which includes:
        (1) Frame quality assessment: The DeQA model is employed to evaluate perceptual clarity and retain visually high-quality frames.  
        (2) Redundancy removal: DINO and FG-CLIP features are extracted from all keyframes to compute cross-frame similarity, filtering out visually redundant samples.  
        (3) Text quality evaluation: A large language model (LLM) scores the ASR-transcribed texts, preserving only high-quality linguistic data.  
\end{itemize}

\textbf{Data Annotation}. The annotation process is organized into two stages, aligned with the different phases of model pre-training. During the first stage, no additional annotations are introduced beyond the automatically extracted keyframes and ASR transcripts. 
In the second stage of pre-training, we incorporate a series of informative annotations to improve convergence efficiency and enhance adaptability to downstream tasks. Specifically:  
(1) Semantic segmentation and summarization: A large language model (LLM)~\cite{yang2025qwen3} is employed to semantically segment and summarize the ASR transcripts, generating coherent textual segments that capture high-level narrative flow.  
(2) Visual captioning: Each scene is annotated using Qwen2.5-VL-7B~\cite{qwen2p5vl}, which produces detailed captions describing the visual content and contextual semantics.  
(3) Multimodal summarization: The LLM~\cite{yang2025qwen3} integrates ASR transcripts, semantic text segments, and visual captions to generate a unified summary for each training sample, providing a compact and semantically rich supervision signal.

\subsubsection{Vision-Text Paired Data}
The vision-text subset consists of approximately 500 million image-text pairs and 30 million video-text pairs. The visual data are primarily derived from the Emu3~\cite{wang2024emu3} training corpus, while the corresponding textual annotations have been re-labeled and enriched using Qwen2.5-VL-7B~\cite{qwen2p5vl} to improve annotation quality, descriptive richness, and alignment accuracy.

We further utilize synthetic image-text pairs generated by state-of-the-art open-source text-to-image (T2I) models~\cite{labs2025flux} to enhance image generation quality, while incorporating recently released open-source vision-language datasets, including InfinityMM~\cite{gu2024infinity} and LLaVA-OV~\cite{llava-onevision}, to strengthen multimodal understanding.
These datasets provide high-quality multimodal annotations with grounded visual references and diverse question-answer formats, thereby enhancing the model’s ability to perform structured reasoning, grounded understanding, and contextually rich responses.

For the video-text pairs, we enhance the Emu3~\cite{wang2024emu3} dataset by applying additional filtering based on motion scores to ensure dynamic visual diversity, and by increasing the frame sampling interval to 1 FPS to balance temporal coverage with computational efficiency.
When multiple clip-text pairs originate from the same video, they are packed sequentially according to their temporal order during training, forming naturally interleaved video-text sequences. This design not only improves training efficiency but also enables the model to better capture long-horizon temporal dependencies and contextual consistency within continuous multimodal data.

\subsubsection{Any-to-Image Data}
The Any-to-Image (X2I) dataset contains approximately 27.35 million samples, compiled from extensive open-source datasets and supplemented with privately constructed in-house data.
The open-source data includes SEED-Data-Edit~\citep{ge2024seed}, WeatherStream~\citep{zhang2023weatherstream}, PromptFix~\citep{yu2024promptfix}, OmniGen-X2I~\citep{xiao2025omnigen}, ShareGPT-4o-Image~\citep{chen2025sharegpt}, ImgEdit~\citep{ye2025imgedit}, OmniGen2-X2I2~\citep{wu2025omnigen2}, MultiRef~\citep{Chen2025MultiRefCI}, GPT-IMAGE-EDIT-1.5M~\citep{wang2025gpt}, and so on.
However, open-source data frequently exhibits inherent limitations, including insufficient diversity, suboptimal quality, and constrained quantity.
To address these limitations, we have curated additional large-scale X2I data for training from a wide array of videos and images, significantly enhancing diversity, quality, and scale.

\subsubsection{Text-only Data}
We integrate a large-scale text-only corpus containing approximately 3 trillion tokens. Building on the text data used in Emu3~\cite{wang2024emu3}, we further expand the dataset by incorporating carefully filtered, high-quality open-source corpora~\cite{li2025infinityinstructscalinginstruction,su2024nemotron} in both English and Chinese, ensuring balanced coverage across languages and domains.
This text corpus establishes a robust foundation for language modeling, preserving strong linguistic capabilities while enhancing the efficiency and generalization of multimodal learning.
By grounding multimodal training in rich and diverse textual knowledge, it enables \Ours\ to produce semantically coherent and logically consistent generations in interleaved vision-language contexts.

\subsection{Training Details}

\textbf{Training Objective.} We adopt the same set of special tokens and multimodal data formatting strategy as Emu3~\cite{wang2024emu3}, integrating visual and textual tokens into unified document-like sequences for pre-training. Since all visual signals in \Ours are fully tokenized into discrete representations, the model is trained using a standard next-token prediction objective based on the cross-entropy loss. To maintain balanced optimization between modalities and prevent visual tokens from overwhelming the training dynamics, a weighting factor of 0.5 is applied to the loss terms corresponding to visual tokens.

\textbf{Training Stage.} Table~\ref{tab:pretraining_recipe} presents the overall training pipeline, covering stage configurations, parallelism strategies, optimization settings, and training procedures. The \Ours\ model is pre-trained through a two-stage process.

\begin{itemize}
\item \textbf{Stage 1 (S1):} The model is pre-trained on 10 trillion tokens, with a maximum sequence length of 32,768 tokens. This stage focuses on large-scale general training, aiming to learn fundamental multimodal alignment and next-token prediction across both visual and textual modalities. 
\item \textbf{Stage 2 (S2):} The model continues pre-training on approximately 3 trillion tokens. This stage further enhances the model’s multimodal generation ability by increasing image resolution, improving data quality, balancing data distribution, and incorporating more interleaved multimodal annotations. 

\end{itemize}

The training and inference infrastructure is built upon the FlagScale~\cite{flagscale2025} framework, which provides comprehensive support for various parallelism strategies, efficient configuration management, and distributed deployment across heterogeneous hardware architectures. Both training stages adopt tensor parallelism (TP) = 8 and context parallelism (CP) = 2. The model is initialized from Qwen3~\cite{yang2025qwen3}. During the first stage (S1), all data are packed online to the maximum context length, enabling efficient utilization of computational resources. All images are constrained to a maximum of 1,024 visual tokens, corresponding to a pixel area of up to $512\times512$ while preserving the original aspect ratio.
In the second stage (S2), the interleaved data augmented with additional annotations are pre-packed offline and padded to the maximum context length to ensure balanced training efficiency and annotation consistency.
A dynamic token strategy is adopted in this stage, with visual token counts ranging from 1,024 to 4,096. Specifically, images are resized to maintain their original aspect ratio, with the minimum resolution set to $512\times512$ and the maximum resolution up to $1024\times1024$.
Throughout all stages, the AdamW optimizer is employed with $\beta_1 = 0.9$, $\beta_2 = 0.95$ and $\epsilon=1.0 \times 10^{-8}$.

\begin{figure}[htbp]
    \centering
    \begin{subfigure}[b]{0.48\linewidth}
        \centering
        \includegraphics[width=\textwidth]{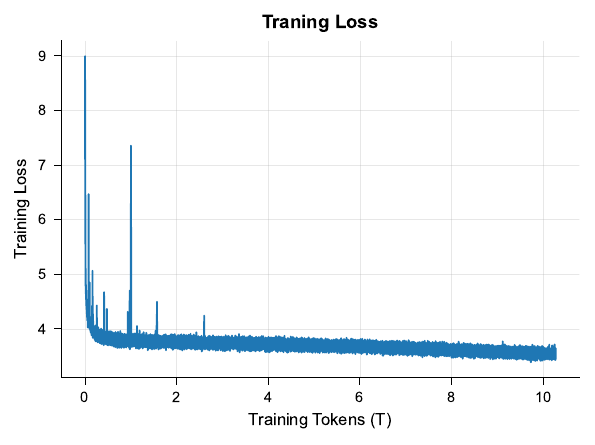}
        \caption{Training loss curve of \Ours during the first stage of pre-training}
        \label{fig:sub1}
    \end{subfigure}
    \hfill
    \begin{subfigure}[b]{0.48\linewidth}
        \centering
        \includegraphics[width=\textwidth]{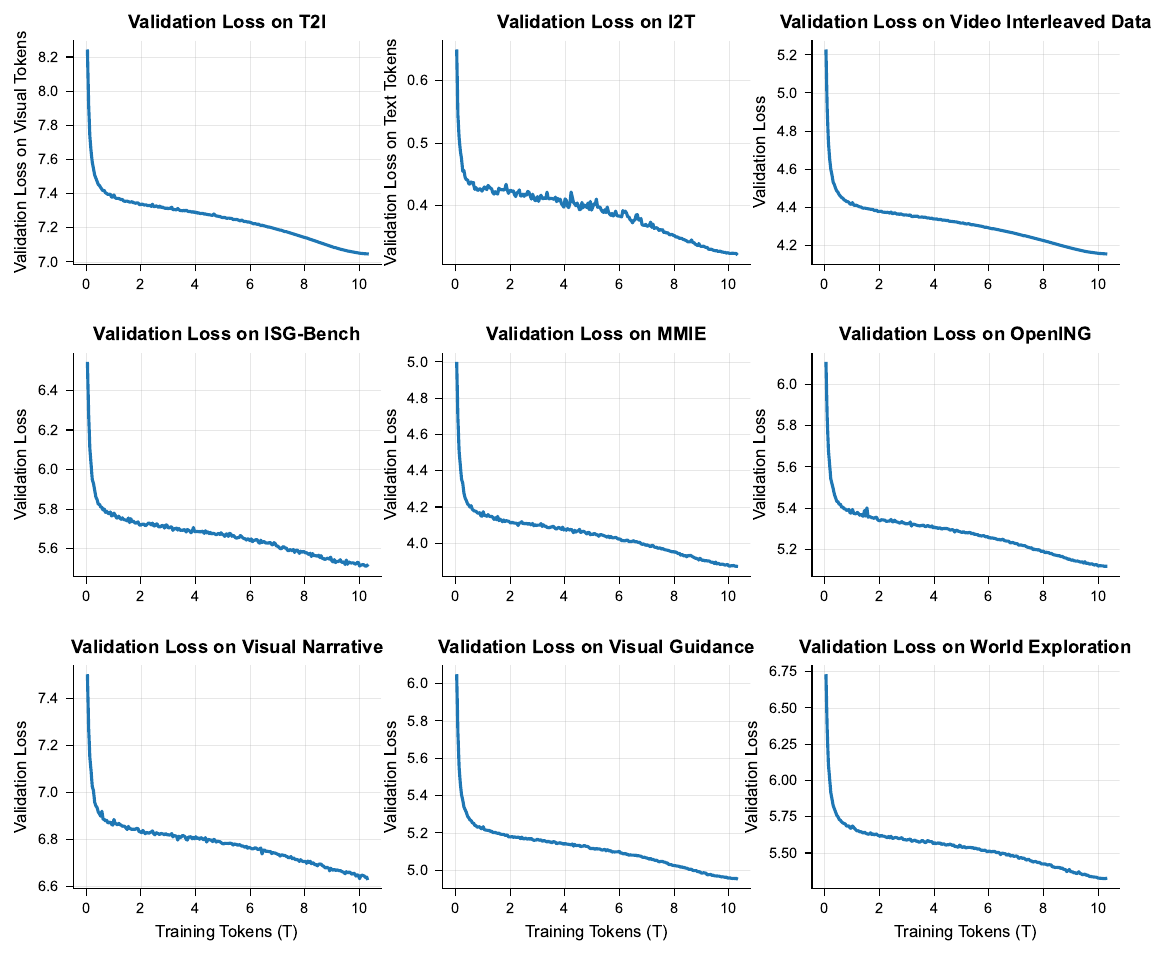}
        \caption{Validation loss curve of \Ours on 9 validation sets during the first stage of pre-training}
        \label{fig:sub2}
    \end{subfigure}
    \caption{Training and validation loss trends of \Ours during the first stage of pre-training. The curves indicate that \Ours achieves smooth and stable optimization, while maintaining consistent generalization across multiple validation datasets.}
    \label{fig:pt-loss}
\end{figure}

Figure~\ref{fig:pt-loss} illustrates the overall optimization dynamics of \Ours during the first stage of pre-training. 
The training loss exhibits a smooth and consistent decline, indicating stable convergence under large-scale multimodal optimization. Similarly, the validation loss across all nine held-out validation sets shows a steady downward trend, reflecting the model’s strong generalization ability across both in-domain and out-of-distribution (OOD) scenarios.

\begin{table}[htbp]
\centering
\begin{tabular}{l|cc}
\toprule
\textbf{Hyperparameters} & \textbf{Stage1} & \textbf{Stage2} \\
\hline
Learning rate  & $5\times10^{-4}$ & $1\times10^{-5}$  \\
LR scheduler   & \multicolumn{2}{c}{Cosine} \\
Weight decay   & \multicolumn{2}{c}{0.1} \\
Gradient norm clip  & \multicolumn{2}{c}{5.0}\\
Loss weight (visual : text)  & \multicolumn{2}{c}{0.5 : 1.0}\\
Warm-up steps   & \multicolumn{2}{c}{700}\\
Training steps  & 700k & 240k  \\
Sequence length & \multicolumn{2}{c}{32768}\\
Batch Size & \multicolumn{2}{c}{448}\\
Resolution     & [512, 512] & [512, 1024] \\
Training seen tokens & 10.3T & 3.5T \\

\midrule
\textbf{Data sampling ratio} & \textbf{Stage1} & \textbf{Stage2} \\
\midrule
Text                            & 0.2   & 0.18   \\
Image-text pair                 & 0.2   & 0.16  \\
Video-text pair                 & 0.05  & 0.08 \\
Any-to-image Data               & 0.0   & 0.03  \\
Video interleaved data          & 0.55  & 0.55  \\

\bottomrule
\end{tabular}
\vspace{0.5em}
\caption{
    Training recipe for \Ours pre-training.
}
\vspace{-5pt}
\label{tab:pretraining_recipe}
\end{table}

The nine validation sets cover a comprehensive range of evaluation perspectives. 
For ISG-Bench~\cite{chen2024interleaved}, OpenING~\cite{zhou2025opening}, and MMIE~\cite{xia2024mmie}, we construct validation samples by concatenating each benchmark question with its ground-truth answer to form coherent input–output pairs for validation loss computation.
Three in-domain validation sets cover the major data types involved in pre-training, including text-to-image (T2I), image-to-text (I2T), and video-interleaved data.
Each set is split from its respective pre-training corpus, where only visual token loss is computed for T2I and only text token loss for I2T, ensuring targeted and modality-specific evaluation. 
The remaining three validation sets are derived from early-stage supervised fine-tuning (SFT) data covering downstream tasks such as visual narrative, visual guidance, and world exploration, with no overlap with the pre-training data.
Consistent improvement across all nine sets confirms that the large-scale interleaved training paradigm yields stable optimization dynamics and robust generalization across modalities and domains.

This first pre-training stage primarily relies on large-scale video interleaved data without introducing any additional annotations, while image-text pairs and text-only data serve as auxiliary sources.
Such a video interleaved data centric scaling paradigm enables stable convergence even with heterogeneous modalities, and more importantly, demonstrates effective generalization across diverse data distributions. 
By leveraging the naturally interleaved structure of video and audio-asr-text pairs, \Ours learns temporal continuity and cross-modal coherence directly from large-scale video data, achieving both scalability and representational robustness.

\section{Post-training}

\subsection{Supervised Fine-tuning}

\subsubsection{Task Formulation}

\textbf{General Tasks.} \Ours naturally supports a broad spectrum of general multimodal tasks, including text-to-image generation (T2I), language understanding and generation (Language), as well as vision-language question answering (VL).

\textbf{Any-to-Image.} As a fundamental capability for multimodal single-step generation and world editing, Any-to-Image (X2I) generation, \ie, general-purpose image editing, is of critical importance. Here, ``X'' denotes arbitrary sequences of interleaved image-text input instructions. 
Various condition-based image generation tasks, such as text-driven image generation, local image editing, subject-driven image generation, in-context image editing, and in-context image generation, all represent specific instantiations or sub-tasks of the broader X2I paradigm.
X2I poses more extensive demands and more complex challenges than conventional image editing and generation tasks, particularly in capabilities and attributes such as multimodal instruction following, subject/background consistency, stylistic and qualitative aspects of generation, world knowledge and physical laws, and so on.
The proficient acquisition of the challenging capabilities required by X2I will facilitate the model's progression towards a more universal Any-to-Any (X2X) generation paradigm, thereby enabling it to address more complex world model tasks.

\textbf{Visual Narrative.}
Visual Narrative, characterized by generating consecutive
storylines with narrative texts and vivid images in an interleaved manner, has emerged as a critical multimodal task with broad practical application. 
This task not only requires the model to generate structure-completed story scripts but also visually engaging images that are temporally consistent in character and style. It poses significant challenges in the higher demand for comprehension of the intricate relationship between visual cues and textual expression to maintain a coherent narrative flow. 
Unlike previous methods~\cite{ruiz2023dreambooth, tewel2024training, avrahami2024chosen, zhou2024storydiffusion} that generate a series of images based on the provided captions or methods~\cite{ge2024seed, shen2025storygpt} that focus on the story with restricted domains, \ie purely cartoon-style stories, our visual narrative drastically expands the frontier towards real-world modeling with two critical capabilities. 
First, our vision-text interleaved story scenario generation encompasses a broad spectrum, ranging from the virtual to the real (\eg, anime, cartoon, daily life occurrences), the ancient to the contemporary (\eg, historical event, movie, vlog), and narrative expression to imaginative creation (\eg, scientific concepts, fairy tale).
Second, our created content demonstrates substantial general knowledge and educational intent, which are presented in an image-text coherent narrative with detailed description and precise and engaging visual depiction. 
Therefore, such ability enables our visual narrative as a bridge for multimodal learning towards holistic world-level comprehension.

\textbf{Visual Guidance.}
Visual Guidance is a multimodal learning task designed to enable models to understand and generate procedural actions through visual information, such as images or video frames. This task requires the model to align visual cues with linguistic expressions across multi-step instructions or operational scenarios, integrating textual commands with concrete visual contexts so that the model can not only comprehend what to do but also how an action should be performed.
Visual Guidance focuses on interleaved vision-language generation, where visual and textual elements are jointly composed to form coherent, step-by-step representations of a process.
In this setting, visual signals are no longer merely auxiliary inputs for recognition or description generation; rather, they function as dynamic guidance that constrains linguistic reasoning and grounds textual instructions within real visual contexts.
By requiring the model to jointly interpret and execute multi-step instructions under both textual and visual conditions, such as cooking, handcrafting, or mechanical assembly, Visual Guidance pushes multimodal learning beyond co-occurrence-level understanding toward action-level comprehension and causal, process-oriented reasoning.
Such capability not only brings multimodal models closer to human-like learning and task execution but also lays a foundation for developing interactive and embodied AI systems that can perceive, reason, and act in the physical world.

\textbf{World Exploration.}
World Exploration is designed to enable models to immerse themselves in user-defined virtual worlds and perform interactive exploration based on textual or multimodal prompts. 
Given a pure text or image-text prompt that specifies the semantic, spatial, or stylistic context of a world, the model produces interleaved vision-language outputs comprising visual observations and corresponding textual narrations, constructing a coherent environment and allowing users to explore it step by step through natural-language instructions or implicit trajectory evolution. 
During exploration, the model must maintain spatial consistency, visual realism, and causal continuity, ensuring that each generated observation accurately reflects both the user’s intent and the evolving world dynamics.
To realize such capabilities, we formulate World Exploration as a unified framework for interactive scene understanding and long-horizon visual synthesis, comprising two complementary paradigms: User-Interactive Mode and Free-Exploration Mode. The User-Interactive Mode focuses on explicit controllability, where each user instruction triggers a single-step visual update corresponding to a deliberate exploration action. In contrast, the Free-Exploration Mode emphasizes autonomous continuity, allowing the model to self-navigate within the initialized environment, producing temporally coherent visual sequences and synchronized textual narrations that describe the unfolding world. The synergy between these two paradigms supports smooth transitions between human-guided and model-driven exploration, balancing controllable interaction with open-ended imagination. This unified design inherently accommodates hybrid real-synthetic environments and dynamic scene evolution, laying the groundwork for embodied reasoning and generative world modeling.

\textbf{Embodied Manipulation.} 
Embodied Manipulation is a fundamental challenge in robotics,
requiring an agent to execute a sequence of dexterous, physical interactions with objects in an environment to achieve a long-term goal. 
Unlike isolated pick-and-place tasks, embodied manipulation is inherently long-horizon, involving multiple intermediate states and a variety of semantic skills (\eg, grasping, pouring, folding). Successfully generating such a manipulation process demands three core capabilities: (1) understanding physical laws and object affordances, (2) planning a correct sequence of subtasks based on the final goal, and (3) generating feasible motions to achieve each intermediate subtask state.
We define the embodied manipulation task in world models by decomposing the long-horizon task into a series of semantically distinct subtasks. Each subtask is represented by a language instruction and a visual keyframe, capturing the essential state change without modeling every instantaneous detail. Formally, given an initial task instruction $L$ and an observation sequence $O = \{o_1, o_2, \cdots, o_T\}$, we decompose the sequence into $N$ subtasks: $Sub_1 = (l_1, O_{[t_0:t_1]}), Sub_2 = (l_2, O_{[t_1:t_2]}), \cdots, Sub_N = (l_N, O_{[t_{N-1}:t_N]})$. Here, $Sub_i$ denotes the $i$-th subtask, $l_i$ is its language instruction, and $O_{[t_{i-1}:t_i]}$ is the corresponding segment of observations. The keyframe for $Sub_i$ is the state $o_{t_i}$ that signifies the subtask's completion.
Therefore, we reframe the problem of long-horizon embodied manipulation as the interleaved prediction of subtask instructions and their corresponding keyframe states.

\subsubsection{Training Data}

\begin{table}[ht]
    \centering
    \setlength{\tabcolsep}{3pt}
    \begin{tabular}{l@{\hskip 6pt}c@{\hskip 18pt}l@{\hskip 8pt}l}
         \toprule
         \textbf{Task} & \textbf{\# Tokens (B)} & \textbf{Question Type} & \textbf{Output Type} \\
         \midrule
         General Tasks           & 29.7 & Language / VL / T2I & Text / Image \\
         Any-to-Image            & 56.2 & Multimodal Generation & Image \\
         Visual Narrative        & 10.1 & Story Generation & Interleaved Narrative \\
         Visual Guidance         & 22.5 & Procedural Reasoning & Interleaved Guidance \\
         World Exploration       & 17.5 & Scene Navigation & Interleaved Scene Synthesis \\
         Embodied Manipulation   & 14.1 & Action Planning & Interleaved Action Plan \\
         \bottomrule
    \end{tabular}
    \vspace{0.5em}
    \caption{
    Summary of tasks and data statistics for \Ours SFT. 
    }
    \vspace{-5pt}
    \label{tab:post_training_data}
\end{table}

\textbf{General Tasks.} For the general multimodal tasks, we curate multiple domain-specific datasets to enhance both task coverage and generation quality.
For the text-to-image generation task, we construct a dataset of approximately 5 million balanced high-quality samples, select through stringent filtering criteria to ensure visual-textual consistency and aesthetic quality. The dataset is deliberately augmented with samples from specialized domains such as cartoons, artistic styles, portraits, and text rendering.
For the language task, we adopt the publicly available Infinity-Instruct dataset~\cite{li2025infinityinstructscalinginstruction}, which contains around 8.9 million high-quality instruction-response pairs, providing comprehensive supervision for language understanding and generation.
For the vision-language question answering task, we utilize the open-source LLaVA-OV~\cite{llava-onevision} dataset, comprising approximately 3.7 million samples, to strengthen the model’s capability in multimodal understanding.

\textbf{Any-to-Image.} 
While it's easy to directly use open-source image editing datasets and distillation datasets from closed-source models, surpassing the performance of these closed-source models remains highly challenging.
This is primarily because such data often suffers from severe deficiencies in quality, diversity, and scale, failing to meet the requirements for training models at various stages. 
Therefore, the key to achieving powerful Any-to-Image (X2I) capabilities lies in comprehensive and effective data research, engineering, and governance.

To ensure that the dataset for Any-to-Image (X2I) exhibits strong diversity, high quality, and sufficient quantity, we extensively consider and construct multiple data collections.
The data format follows a specific structure: the input always encompasses a textual instruction along with an optional image set (containing zero, one, or multiple images), and the output is strictly limited to a single generated image.
The primary entities that are modified within the dataset can be categorized into several types, including, for example, human, animal, still object, text, scene, and composite categories.

To diversify the data sources, the dataset is constructed from three primary categories: fully real, semi-real/semi-synthetic, and fully synthetic.
To construct the fully real data, a pipeline involving techniques such as video scene segmentation, video keyframe matching and extraction, and image retrieval is applied to long videos, short video clips, and web-scale images.
To build the semi-real/semi-synthetic and fully synthetic data, a variety of open-sourced models are applied as auxiliary tools to real and synthetic images.
For quality assurance, the majority of the data is filtered according to metrics including image resolution, clarity, and aesthetic quality. 
Subsequently, image clustering is further applied to generate compact yet diverse dataset subsets.

\textbf{Visual Narrative.}
To construct a high-quality visual narrative dataset, we cover a wide spectrum of content, from imaginative and fictional stories to educational narratives and real-world events (\eg, scientific concepts, fairy tales, historical events, daily life occurrences). Our data sources span diverse domains, enabling the collection of video data with rich visual and textual content. All videos are first processed by scene segmentation\footnote{\url{https://github.com/Breakthrough/PySceneDetect}} and automatic speech recognition (ASR)~\citep{whisper} to obtain structured visual and textual sequences, similar to the pre-training data processing pipeline. 
To transform these sequences into coherent short narratives, we carefully design a multi-step processing framework. We first extract visual features~\citep{DINOv2,FGCLIP} and quality scores~\citep{DEQA} for each keyframe, followed by deduplication based on feature similarity and quality score to produce compact, representative sequences. 
As the deduplicated sequences often contain multiple interleaved narratives and existing VLMs exhibit limited capability in handling multi-image multi-text input prompts, we opt to generate a dense caption for each frame using Qwen2.5-VL~\citep{qwen2p5vl} and concatenate them with ASR transcripts as input to Qwen3~\citep{yang2025qwen3} for accurate short narrative segmentation, which substantially improves boundary precision.
Each segmented story is then evaluated by Qwen3 for narrative completeness, ensuring consistency in content and characters while maintaining a coherent story arc with well-defined beginnings, developments, and conclusions.
The verified stories are further refined by filtering keyframes and constructing narrative text prompts based on ASR transcripts and visual content.
Finally, three types of reasoning-oriented annotations are independently generated for each narrative, including questions(user prompts), global chain-of-thought (CoT), and image-level CoTs.
Through this pipeline, we construct a coherent, high-quality visual narrative dataset containing a total of 430k samples in both Chinese and English, enabling multimodal models to advance in visual narrative generation and reasoning.

\textbf{Visual Guidance.}
To construct a large-scale interleaved visual-text dataset for the Visual Guidance task, we collect diverse real-world instructional data encompassing multimodal demonstrations, step-by-step tutorials, and procedural guides across everyday scenarios such as cooking, DIY, and handcrafting. Using task-related keywords and metadata, we retrieve high-quality instructional videos and extract key textual segments from subtitles, aligning them with representative video keyframes to form coherent step-wise image-text pairs that capture procedural actions.
Before refinement, we remove corrupted or low-quality samples, including those with distorted or unbalanced aspect ratios, duplicated or semantically redundant text, and format inconsistencies. Samples with fewer than two or more than ten procedural steps are also filtered out.
Inspired by recent reasoning models, we further introduce a dual-level Chain-of-Thought (CoT) mechanism to enhance multimodal coherence. The image-level CoT leverages Qwen3~\cite{yang2025qwen3} and Qwen2.5-VL~\cite{qwen2p5vl} to derive detailed visual reasoning for each step, benefiting from the models’ strong image generation and editing capabilities. 
The global CoT, on the other hand, provides a global semantic layout that maintains logical consistency and mitigates long-sequence forgetting.
Finally, Qwen2.5-VL is employed to automatically score and rank each sample along multiple dimensions—including step relevance, instructional clarity, image-text alignment, and visual informativeness. After comprehensive filtering and quality control, we obtain a final corpus of 960K high-quality interleaved samples, containing both Chinese and English instructional data with strong procedural, temporal, and semantic grounding for multimodal learning.

\textbf{World Exploration.}
To construct a large-scale and high-quality dataset for the World Exploration task, we build upon the open-source Sekai~\citep{li2025sekai} and OpenDV~\citep{yang2024generalized} corpora, which together provide complementary walking and driving exploration scenarios covering both real-world and game-engine environments. 
This combination ensures diverse spatial layouts, motion dynamics, and scene semantics, providing a rich foundation for models to learn open-ended environment construction, spatial reasoning, and continuous exploration under both user-guided and autonomous settings. 
Based on the above diverse data sources, we begin by applying the DeQA-Score filtering scheme~\citep{DEQA} to remove low-quality or visually degraded video clips, ensuring high fidelity and stable frame-level coherence. 
Following the camera-pose annotation pipeline proposed in~\citep{leroy2024grounding}, we re-annotate all retained clips with precise camera trajectories, guaranteeing reliable spatial alignment and viewpoint consistency across frames.
To enhance multimodal reasoning quality, we further employ Qwen3~\citep{yang2025qwen3} and Qwen2.5-VL~\citep{qwen2p5vl} to automatically generate user prompts and image-level CoT annotations for each sample, capturing causal motion reasoning and perception-action alignment. 
In this manner, each annotated clip is paired with fine-grained exploration directives, describing explicit viewpoint transitions, motion intents, and semantic focus shifts along the trajectory. 
Based on the obtained exploration directives, each sample is transformed into four interleaved instances combining input modalities and interaction modes: two input modalities, pure-text prompts containing only linguistic instructions and multimodal prompts where the first keyframe provides contextual initialization; and two exploration paradigms, User-Interactive Mode supporting stepwise instruction–response interactions and Free-Exploration Mode modeling autonomous continuous traversal with temporally coherent visual and textual outputs.
After comprehensive filtering, re-annotation, and reasoning enrichment, we obtain a total of 200K high-quality World Exploration samples, each represented as an interleaved sequence of visual frames, textual observations, and exploration directives. This dataset provides a strong foundation for training and evaluating models capable of spatiotemporal consistency, embodied reasoning, and open-ended world interaction.

\textbf{Embodied Manipulation.}
To construct a large-scale dataset for interleaved subtask-keyframe prediction in Embodied Manipulation, we integrate and process data from three primary sources: the OpenX Embodiment dataset (OXE~\citep{o2024open}), the Agi-world Alpha dataset~\citep{agibot}, and a self-collected Songling Aloha dataset. Our final dataset comprises 973K samples, each segmented into semantic subtasks annotated with keyframes and descriptions.
The processing pipeline differs based on the source dataset. For the Songling Aloha dataset, we employ an online labeling system, similar to the one used for Agi-world Alpha, to manually partition each trajectory into contiguous clips corresponding to distinct subtask stages. 
For the extensive and diverse OpenX Embodiment dataset (containing approximately 1 million episodes), we develop an automated labeling framework. This framework first applies a keyframe candidate selection algorithm to identify frames with significant motion or gripper state changes. Subsequently, we utilize Qwen2.5-VL~\citep{qwen2p5vl} to merge these keyframes into adjacent segments and generate a descriptive subtitle for each. The generated segments undergo a final quality control step, where low-quality segments are filtered out, and adjacent segments with semantically similar descriptions are merged.
To enhance the model's capability for recovery and multi-step planning, we construct sequences not only from the initial state but also by randomly starting from intermediate steps, forcing the model to predict successive subtasks and keyframes from any arbitrary point in the trajectory.
In summary, our interleaved vision-language dataset consists of approximately 973K samples: 920K from OXE, 40K from Agi-world Alpha, and 13K from Songling Aloha.

\subsubsection{Training Details}

After the pre-training phase, we integrate high-quality data from various multimodal tasks and perform unified supervised fine-tuning (SFT) to establish a shared multimodal interface, facilitating mutual enhancement and knowledge transfer across different downstream tasks.
To further enhance the model's performance and training efficiency while ensuring high-resolution generation quality, we employ a two-stage SFT strategy. The detailed data statistics for the downstream tasks are shown in Table~\ref{tab:post_training_data}.

In the first stage, we train the model on each downstream task at a standard resolution. Specifically, the Any-to-Image task uses a 768px resolution, while Visual Guidance, Visual Narrative, and Embodied Manipulation tasks are trained at 512px. The World Exploration task, requiring more detail, is trained at a 720px resolution. For the visual modality, we set a weight factor of 1.0, applied to the loss terms corresponding to all visual tokens. The maximum sequence length during training is set to 16,384 to balance performance and computational load.
In the second stage, we further train the model at higher resolutions. The Any-to-Image task is trained at 1024px, and other interleaved tasks are extended to 720px. As more visual tokens are introduced, we set the weight factor for visual tokens to 0.5 to maintain balanced optimization between modalities. 
Additionally, we expand the maximum sequence length to 32,768. This stage improves generation quality, particularly in high-resolution image generation and the accuracy of cross-modal task execution, further promoting knowledge transfer between tasks.

Consistent with the pre-training phase, the training and inference infrastructure for the SFT phase is built upon the FlagScale~\cite{flagscale2025} framework. In the first stage, Tensor Parallelism (TP) is set to 8, and Context Parallelism (CP) is set to 1. In the second stage, TP remains at 8, and CP is increased to 2. We set the batch size to 1024 and a learning rate of $6e-6$. Throughout all stages, the AdamW optimizer is employed with $\beta_1 = 0.9$ and $\beta_2 = 0.95$, along with a cosine learning rate schedule. For general tasks such as text-to-image generation, language understanding and generation, as well as vision-language question answering, we pre-pack and pad the input data in both stages to the maximum context length to ensure balanced training efficiency and data consistency. Each stage is trained for 3000 iterations, allowing the model to progressively adapt to the specific characteristics of the tasks and optimize across different modalities.

\subsection{Reinforcement Learning}

\subsubsection{Reward System}
To enhance multimodal reasoning and generation, \Ours is further post-trained with large-scale joint reinforcement learning training across multiple multimodal tasks for the first time. To enable this, we construct a comprehensive reward system composed of multiple distinct rewards, which provides thorough and unified guidance for diverse downstream tasks. Our reward system features three essential characteristics: 

\textbf{(i) Generality:} We design general reward components to offer universal guidance and generate rewards applicable to general generation tasks, such as aesthetic quality and image-text alignment. In practice, these general components include CLIP-based image-text similarity, VLM-based alignment accuracy, and aesthetic scorers for overall visual appeal. These universal signals provide shared optimization objectives that are valid across text, image, and interleaved modalities.

\textbf{(ii) Task-specificity:} We decouple rewards to provide task-exclusive guidance to address task-specific challenges. For instance, OCR- and layout-based text fidelity scoring is employed for text rendering tasks, face-detection and identification are used for human identity preservation in X2I task, and VLM-based consistency metrics are used for narrative and reasoning-oriented tasks. This modular design allows each task to retain a unique optimization signal while still be adaptable to the unified reward framework.

\textbf{(iii) Unified nature:} We perform end-to-end training within a single reward space. By jointly leveraging reward signals from diverse downstream tasks, we guide the reinforcement learning optimization process in a unified manner, ensuring that different objectives complement rather than interfere with one another. To balance the heterogeneous reward distributions across tasks, each reward signal is normalized to the range [1, 10] before being combined, maintaining consistent optimization scales across all objectives.

This multi-dimensional reward system ensures that \Ours balances various quality criteria simultaneously, and more importantly, naturally avoids hacking into a single reward, thereby achieving consistent improvements across multiple tasks without backfiring the performance of individual tasks.

\subsubsection{Training Data}
The reinforcement learning phase is conducted on curated subsets derived from the supervised fine-tuning (SFT) data, supplemented by user feedback and targeted data construction for image-centric tasks. For each downstream task, we filter approximately {10K} high-quality prompts with diverse contexts, and incorporate an additional {1K} human feedback samples to better align the optimization with human preference. Furthermore, we collect extra {58K} high-quality and diverse X2I instructions and {50K} T2I samples to support the X2I-focused reinforcement learning stage.

\subsubsection{Training Details}
\Ours is trained in a unified multi-task, multi-modal reinforcement learning setup with a distributed reward system for efficient large-scale feedback. \Ours is optimized using the Group Relative Policy Optimization (GRPO) algorithm~\cite{shao2024deepseekmath}, with a global batch size of 640, a learning rate of $1\times10^{-6}$, and a rollout number of 8. Rollouts are performed using a vLLM-based sampling engine integrated into the VeRL framework~\cite{sheng2024hybridflow} for stable and scalable generation. The unified multi-task reinforcement learning phase completes one full pass over all collected prompts, combining tasks within each batch to encourage cross-task synergy.
To improve single-image generation quality and editing consistency, we apply a separate reinforcement learning stage using X2I and T2I data. Unless otherwise specified, all reported quantitative results are based on the resulting RL model.

\begin{figure}[t]
    \vspace{-15pt}
    \centering
    \includegraphics[width=0.8\linewidth]{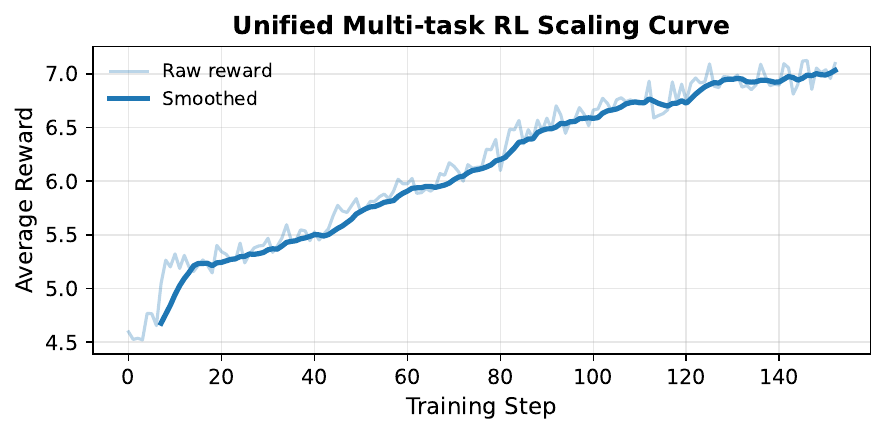}
    \caption{Average reward steadily increases from $\sim$4.5 to $>7.1$ during multi-task RL training.}
    \label{fig:rl_scaling_curve}
\end{figure}

\subsubsection{Scaling Behavior}

We monitor the evolution of the averaged reward score throughout the entire mixed-task RL training process. As shown in Figure~\ref{fig:rl_scaling_curve}, the curve exhibits a consistently increasing trend from an initial reward of around {4.5} to over {7.1}, demonstrating stable and continuous improvement.

The reward curve indicates that \Ours successfully balances multiple heterogeneous objectives within a single unified optimization process. This scaling pattern suggests that the unified reward aggregation mechanism effectively integrates task-specific feedback while preserving general multimodal consistency, confirming the scalability and robustness of our reinforcement learning design.

\subsection{\Ursa}

\subsubsection{Training Approach}

\begin{figure}[htbp]
    \centering
    \includegraphics[width=0.9\linewidth]{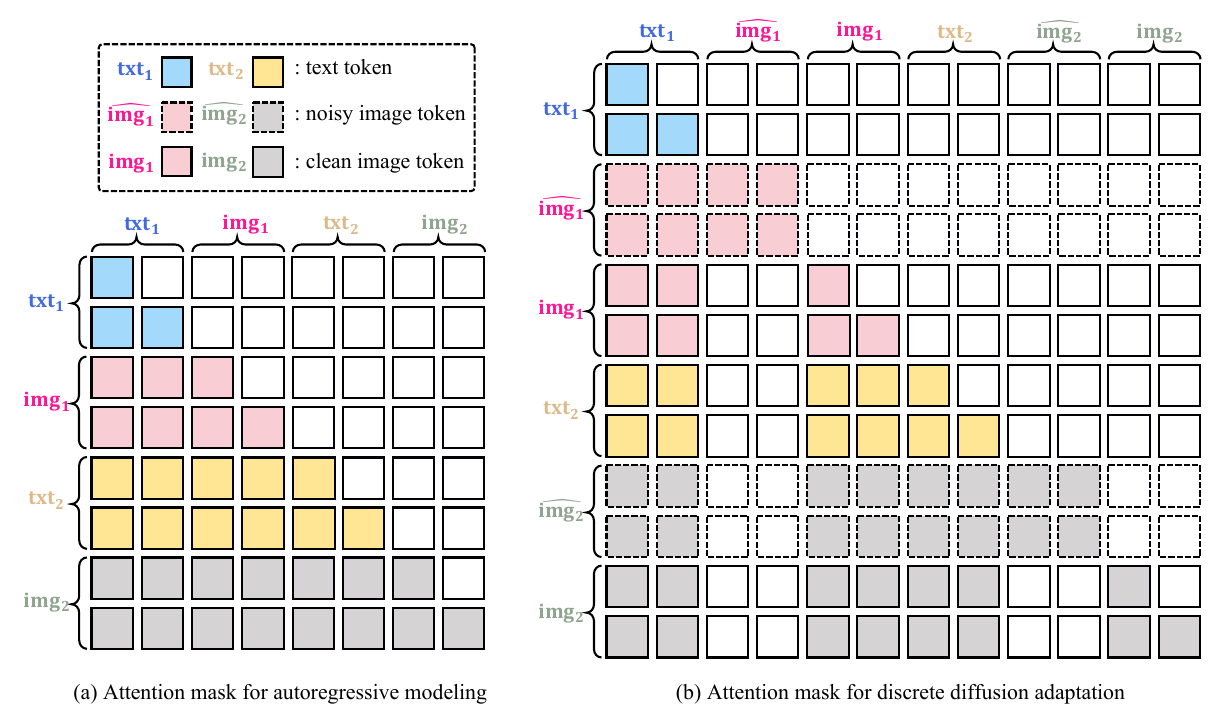}
     \caption{\textbf{\Ursa.} \textbf{(a)} The model performs standard next-token prediction for large-scale multimodal pre-training, supervised fine-tuning, and reinforcement learning. \textbf{(b)} During discrete diffusion adaptation, each image is duplicated with a noisy copy. Noisy tokens attend causally to preceding clean tokens and bidirectionally to noisy tokens within the same image, while clean image and text tokens follow the original causal pattern to preceding clean tokens.}
    \label{fig:dfm_attn_mask}
\end{figure}

Despite the strong generative capability, multimodal autoregressive models are inherently hindered by token-by-token sequential decoding, leading to low inference efficiency, particularly in image generation. For instance, generating a $1024\times1024$ image with $16\times$ downsampling ratio requires roughly 4K tokens, resulting in considerable computational latency.

To address this, we propose \Ursa (\Ursaabbr), a lightweight adaptation approach that accelerates autoregressive image generation while leaving the model’s text generation capabilities unchanged. Built upon a pre-trained autoregerssive model, \Ursaabbr extends the discrete diffusion~\citep{DFM:DFM, ursa} formulation to visual tokens, allowing the model to transition image generation from sequential decoding to parallel generation. Concretely, \Ursaabbr implements a discrete diffusion process over visual tokens, where the entire image token sequence is initialized at once and progressively refined through a series of discrete denoising steps to recover the target image. This formulation enables significantly faster inference without sacrificing output quality.

For training, we construct a self-distillation dataset of image-text pairs and interleaved image-text sequences. To accommodate discrete diffusion training with the interleaved text-image sequence, we modify the attention masks to enable global modeling of visual tokens while preserving accurate text-visual relationships. 
Specifically, as illustrated in Figure~\ref{fig:dfm_attn_mask}, each noisy image token attends causally to preceding clean tokens, while attending bidirectionally to other noisy tokens within the same image. In contrast, each clean image or text token follows the original causal scheme, attending only to preceding clean tokens.

\subsubsection{Infrastructure}

While existing infrastructures~\citep{megatron, vllm} provide solid foundations, they fall short in flexible cross-modal attention training and dynamic modality-switching inference needed for \Ursaabbr.
Built upon FlagScale~\citep{flagscale2025}, we introduce several key innovations to address these challenges in both training and inference.

\textbf{Flexible Cross-Modal Attention and Hybrid Parallel Training.} To efficiently model complex cross-modal attention patterns inherent in architectures such as \Ursaabbr, we extend the FlagScale framework by integrating PyTorch FlexAttention. Instead of the conventional 4D attention mask, we adopt a per-row block mask that flexibly encodes causal, bidirectional, and region-specific attention constraints. This design eliminates the need to store the full attention matrix, significantly reducing memory consumption and enhancing scalability for long sequences. To further optimize distributed training efficiency, we adopt a hybrid parallelism strategy combining Tensor Parallelism (TP), Pipeline Parallelism (PP), Sequence Parallelism (SP), and ZeRO-1 Data Parallelism (DP). In addition, activation recomputation is applied to minimize the memory footprint while maintaining training stability. 

\begin{figure}[h]
    \centering
    \adjustbox{max width=1.10\textwidth, center}{
        \includegraphics[width=1.2\textwidth]{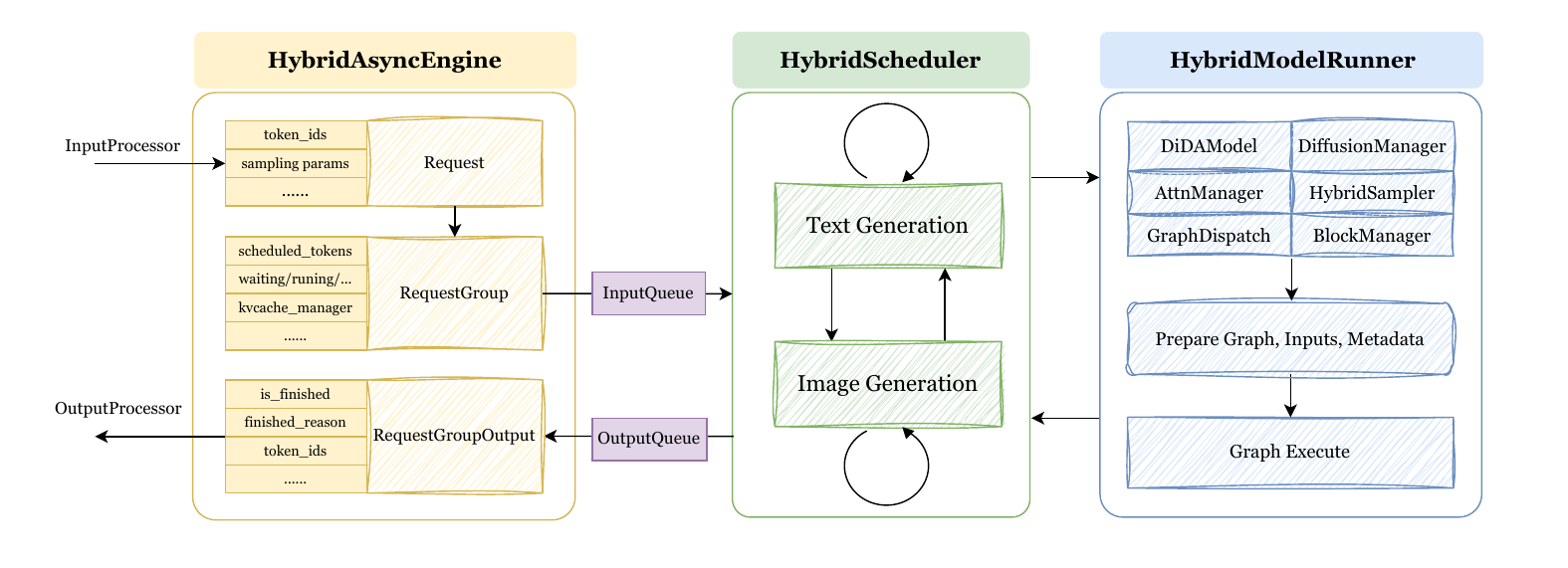}
    }
    \vspace{-20pt}
    \caption{Hybrid Inference Framework with FSM-based Scheduling.}
    \vspace{-5pt}
    \label{fig:dida_infer}
\end{figure}

\textbf{Hybrid Inference Framework with FSM-based Scheduling.\!\!\!} To support complex dynamic modality-switching inference, we developed a hybrid inference framework for the \Ursaabbr model within FlagScale, as illustrated in Figure \ref{fig:dida_infer}. 
Specifically, we introduces a Finite-State Machine (FSM)-based scheduler that adaptively manages transitions between text and image phases while preallocating resources, enabling efficient concurrent processing. 
Combined with asynchronous request handling, runtime state reuse, and FP8 quantization, the framework substantially reduces kernel overhead and increases throughput, achieving at least 50\% speedup on a 4-device setup.

\section{Tokenizer Training}
\subsection{Data}
\paragraph{Image Corpus.\!\!\!}
To fully excavate the potential of the representational capacity of the visual tokenizer, we curate a large-scale training data with versatile domains. (1) General: ImageNet~\cite{imagenet}, OpenImage~\cite{openimage}, CC3M~\cite{cc3m}, CC12M~\cite{cc12m}, and diverse in-house data that comprises the category of movies, gameplay recording, vlogs, etc., are applied for general distribution learning. (2) Aesthetic: We collect high-quality images from open-source websites to establish the aesthetic dataset. (3) Specific: To ensure the precise modeling of textual and facial representation, we construct a text- and face-rich dataset. TextAtlas5M~\cite{textatlas5m}, PosterCraft~\cite{postercraft}, LAION~\cite{laion5b} are extracted for text-rich dataset curation. While for the facial one, RetinaNet~\cite{lin2017focal} is employed for facial data derivation where it is mainly from Midjourney~\cite{midjourney}, COYO-700M~\cite{kakaobrain2022coyo-700m}, DataComp~\cite{datacomp}, and JourneyDB~\cite{pan2023journeydb}.

The image data are processed similarly: (1) We apply a resolution filter, discarding samples with a resolution below 512 $\times$ 512 pixels. (2) The image quality scoring operator~\cite{yang2022maniqa} provides a synthetic score based on an image’s sharpness, noise level, and clarity. We exclude images with scores below $0.4$ to ensure the overall quality. (3) We employ LAION-AI aesthetic predictor\footnote{\url{https://github.com/LAION-AI/aesthetic-predictor}} for image aesthetics assessment, filtering out samples with low aesthetic scores. (4) The watermark detector is leveraged to remove the images that include any watermark.

\paragraph{Video Corpus.\!\!\!}
The video decoder is trained on a comprehensive mixture of datasets that capture a wide range of visual dynamics and semantic domains. Specifically, we use Koala~\citep{koala36m} for general real-world scenes, Sekai~\cite{li2025sekai} for freeform world exploration, and Agibot~\cite{agibot} for robotic manipulation tasks. We also include a high-quality in-house stock video collection, which offers visually appealing clips, further enhancing the overall fidelity and robustness of the model. In addition, we also include a subset of high-quality image data during training.

We apply the following filtering strategies to extract the high-quality portions of the videos. (1) We use PySceneDetect\footnote{\url{https://github.com/Breakthrough/PySceneDetect}} to segment videos into individual scenes. (2) We compute optical flow~\citep{raft} to remove clips with either minimal or excessively large motion. (3) For Koala data, we apply the Video Training Suitability Score (VTSS)~\citep{koala36m} to filter out samples suitable for training.

\subsection{Training Details}

\paragraph{Tokenizer.\!\!\!}
Following IBQ~\cite{VQ:IBQ}, we train our tokenizer with weighted balance objectives, \ie, reconstruction loss, quantization loss, perceptual loss from LPIPS~\cite{lpips}, adversarial loss with PatchGAN~\cite{patchgan} as the discriminator, entropy loss, and semantic distillation loss with siglip-large-patch16-256~\cite{SigLIP}. We employ Adam as the optimizer with $\beta_{1} = 0.5$ and $\beta_{2}=0.9$. The learning rate is set to $1e-4$ with 15K warmup steps and the overall training iteration is 500K with a total batch size of 256.

\paragraph{Image Decoder.\!\!\!}
For an optional alternative to the vanilla image decoder, we train a diffusion-based image decoder with Flow-matching~\cite{flowmatching, sit} as the training objective and Stable-Diffusion 3.5 medium~\cite{esser2024scaling} as the initialization.
To ensure our image decoder can process multiple resolutions with different aspect ratios, we incorporate a two-stage training strategy. In the first stage, the image decoder is pre-trained on a single 512px for 200k steps with $512$ batch size and a $5e-5$ learning rate until convergence. It helps to adapt the fine-grained  inputs, \ie, low-level guidance from quantized tokens, as the condition. Then, we apply a balanced bucket sampling strategy for 50K iterations fine-tuning, allowing the image decoder to process input images with varying aspect ratios of multiple resolutions ranging from 512px to 1024px. Subsequently, we adopt 6K steps with a total batch size of 128, the learning rate of $1e-5$ for LoRA-based distillation, which decreases the denoising steps from 50 to 4. 
Therefore, \Ours supports different images as input and generates high-fidelity outputs up to 2K resolution with fast speed.

\paragraph{Video Decoder.\!\!\!}
We initialize the video decoder with Wan2.2 5B~\citep{wan2p1} and adopt a progressive training strategy. Specifically, we first pre-train on 720px/480px 1-second clips for approximately 80K steps, using a batch size of 1,024, a learning rate of $5e-5$, and a data composition ratio of $4:1$ between images and videos, to establish strong reconstruction ability and basic motion understanding. Subsequently, we perform hybrid training on clips ranging from 2 to 5 seconds, enabling the model to generalize across varying keyframe intervals. This stage employs dynamic bucket sampling for another 80K steps, with a balanced image-video ratio ($1:1$) and a learning rate of $3e-5$. Finally, we fine-tune the model on 1080px data to enhance high-resolution video generation quality.

\section{Experiment}

\subsection{Text-to-Image}

We assess \Ours's performance in the text-to-image (T2I) task through two dimensions: general generation capacity and text rendering ability. To gauge the model’s general generation performance, we carry out evaluations across four publicly accessible benchmarks: GenEval~\cite{ghosh2024geneval}, DPG-bench~\cite{hu2024ella}, OneIG-Bench~\cite{chang2025oneig} and TIIF~\cite{wei2025tiifbenchdoest2imodel}. These benchmarks offer a thorough assessment of the model’s capacity to produce high-quality images that align semantically with given text prompts. 

To evaluate the model’s text rendering capability, we conduct evaluation both on English and Chinese text generation. For English text rendering, we utilize the LeX-Bench~\cite{zhao2025lex}, CVTG-2K~\cite{du2025textcrafter} benchmark to test the readability of rendered English text. For Chinese text rendering, we performed an evaluation using LongText-Bench~\cite{geng2025x}. This benchmark is designed to assess how well models render longer texts in both English and Chinese.

\textbf{Quantitative Evaluation.} (1) \textbf{TIIF:} Table~\ref{tab:tiif} presents a comparative analysis of model performance on TIIF-Bench mini~\cite{wei2025tiifbenchdoest2imodel}, a specialized benchmark developed to systematically assess the capability of T2I models in interpreting and adhering to complex textual instructions. Notably, \Ours achieves the best avg score illustrating its robust competence in following the instructions.
(2) \textbf{OneIG-Bench:} Table~\ref{tab:oneig_en} and Table~\ref{tab:oneig_zh} report \Ours's performance on OneIG-Bench~\cite{chang2025oneig} which is a comprehensive benchmark for fine-grained evaluation of T2I models across diverse dimensions. \Ours achieves the superior performance in the English track and secures the second position in the Chinese track, which indicates its strong general T2I generation capabilities.
(3) \textbf{LeX-Bench:} Table~\ref{tab:lexbench} reports the quantitative results of English rendering on LeX-Bench~\cite{zhao2025lex}. This benchmark comprises of 1,310 carefully designed prompts for robust text accuracy evaluation, covering diverse fonts and styles. \Ours outperforms all open-source and closed-source models, and achieves significant improvements particularly in the hard category. This demonstrates the promising English text rendering capability of \Ours.
(4) \textbf{LongText-Bench:} Table~\ref{tab:longtextbench} presents quantitative evaluation results for models on LongText-Bench\cite{geng2025x}, a dedicated benchmark constructed to assess the model’s proficiency in precisely rendering long textual content. As shown in the table, \Ours achieves the highest result in rendering long English text and occupy the second-highest place in rendering long Chinese text. These findings collectively demonstrate \Ours’s superior capability in processing and rendering lengthy textual inputs.
(5) \textbf{CVTG-2K:} Table~\ref{tab:cvtg2k} presents quantitative results for English text rendering on the CVTG-2K~\cite{du2025textcrafter}. This benchmark comprises 2,000 prompts, where each prompt requires 2 to 5 distinct text regions rendering on the generated image in English. For evaluation, two specific metrics are employed: Word Accuracy and Normalized Edit Distance (NED). As illustrated in the table, \Ours outperforms state-of-the-art T2I models by a large margin, further demonstrating its strong capability in English text rendering.
(6) Table~\ref{tab:model_comparison_t2i} compares the text-to-image generation performance between Emu3 and \Ours in terms of GenEval~\cite{ghosh2024geneval} and DPG-Bench~\cite{hu2024ella} benchmarks. \Ours generates higher-resolution images ($1024^2$ vs. $720^2$) using nearly half the number of visual tokens, and \Ours surpass Emu3 by a large margin among all metrics and yields promising improvements in both semantic alignment and aesthetic quality. Despite $4 \times$ parameters than Emu3, \Ours maintains comparable inference efficiency in generating the image with the same resolution.

\begin{table}[htbp]\centering
\renewcommand{\arraystretch}{1.7} 
\setlength{\tabcolsep}{3pt}

\centering
\begin{adjustbox}{width=1.0\textwidth}
\begin{tabular}{l|cc|cccccccc|cccccccccccc|cc}
\toprule
\multirow{3}{*}{\textbf{Model}}
  & \multicolumn{2}{c|}{\multirow{2}{*}{\textbf{Overall}}}
  & \multicolumn{8}{c|}{\textbf{Basic Following}}
  & \multicolumn{12}{c|}{\textbf{Advanced Following}}
  & \multicolumn{2}{c}{\textbf{Designer}} \\

\cmidrule(lr){4-11} \cmidrule(lr){12-23} \cmidrule(lr){24-25}

& & &
  \multicolumn{2}{c}{Avg}                    
  & \multicolumn{2}{c}{Attribute}
  & \multicolumn{2}{c}{Relation}
  & \multicolumn{2}{c|}{Reasoning}
  & \multicolumn{2}{c}{Avg}                  
  & \multicolumn{2}{c}{\makecell{Attribute\\+Relation}}
  & \multicolumn{2}{c}{\makecell{Attribute\\+Reasoning}}
  & \multicolumn{2}{c}{\makecell{Relation\\+Reasoning}}
  & \multicolumn{2}{c}{Style}
  & \multicolumn{2}{c|}{Text}
  & \multicolumn{2}{c}{\makecell{Real\\World}} \\

& short & long &         
  short & long &          
  short & long &          
  short & long &          
  short & long &         
  short & long &         
  short & long &        
  short & long &        
  short & long &        
  short & long &        
  short & long &        
  short & long         
\\
\midrule

\multicolumn{25}{l}{\bf{Diffusion based Models}} \\

\midrule

FLUX.1 [dev]~\citep{FLUX}  &{{71.09}} &71.78 &83.12 &78.65& 87.05 & 83.17 & 87.25 &80.39 &75.01 &72.39 &65.79 &68.54& 67.07 &73.69 &73.84 &73.34 &69.09 &71.59 & 66.67 & 66.67 &43.83 &52.83 &70.72 &71.47 \\
FLUX.1 [Pro]~\citep{FLUX} &67.32 &69.89 &79.08 &78.91 &78.83 &81.33 &82.82 &83.82 &75.57 &71.57 &61.10 &65.37 &62.32 &65.57 &69.84 &71.47 &65.96 &67.72 &63.00 &63.00 &35.83 &55.83 &71.80 &68.80 \\
DALL-E 3~\citep{betker2023dalle3} &74.96 &70.81 &78.72 &78.50 &79.50 &79.83 &80.82 &78.82 &75.82 &76.82 &73.39 &67.27 &73.45 &67.20 &72.01 &71.34 &63.59 &60.72 &89.66 &86.67 &66.83 &54.83 &72.93 &60.99 \\
SD 3~\citep{esser2024scaling}            &67.46 &66.09 &78.32 &77.75 &83.33 &79.83 &82.07 &78.82 &71.07 &74.07 &61.46 &59.56 &61.07 &64.07 &68.84 &70.34 &50.96 &57.84 &66.67 &76.67 &59.83 &20.83 &63.23 &67.34 \\
MidJourney v7~\citep{midjourney} &68.74 &65.69 &77.41 &76.00 &77.58 &81.83 &82.07 &76.82 &72.57 &69.32 &64.66 &60.53 &67.20 &62.70 & 81.22 &71.59 &60.72 &64.59 &83.33 &80.00 &24.83 &20.83 &68.83 &63.61 \\
Seedream 3.0~\citep{gao2025seedream} & 86.02 & 84.31 & \underline{87.07} & 84.93 & \underline{90.50} & 90.00 & \textbf{89.85} &85.94 & \underline{80.86} &78.86 & 79.16 &80.60 & 79.76 & 81.82 & 77.23 & 78.85 & 75.64 & 78.64 & \textbf{100.00} & \underline{93.33} & \underline{97.17} &87.78 & 83.21 &83.58 \\
GPT Image 1~\citep{openai_image_api} &\underline{89.15} &\textbf{88.29} &\textbf{90.75} &\textbf{89.66} &\textbf{91.33} &87.08 &84.57&84.57 &\textbf{96.32} &\textbf{97.32} &\textbf{88.55} &\textbf{88.35} &\textbf{87.07} &\textbf{89.44} &\textbf{87.22} &\underline{ 83.96} &\textbf{85.59} &\textbf{83.21} & \underline{90.00} & \underline{93.33} &89.83 &86.83 &89.73 &\textbf{93.46} \\
Qwen-Image~\cite{wu2025qwen} & 86.14 & 86.83 & 86.18 & 87.22 & \underline{90.50} & \underline{91.50} & 88.22 & \textbf{90.78} & 79.81 & 79.38 & 79.30 & 80.88 & 79.21 &78.94 & 78.85 &81.69 & 75.57 &78.59 & \textbf{100.00}  &\textbf{100.00} & 92.76 & \underline{89.14} & \underline{90.30} & 91.42 \\

\midrule   

\multicolumn{25}{l}{\bf{AR based Models}} \\

\midrule

Show-o~\citep{show-o} &59.72 &58.86 &73.08 &75.83 &74.83 &79.83 &78.82 &78.32 &65.57 &69.32 &53.67 &50.38 &60.95 &56.82 &68.59 &68.96 &66.46 &56.22 &63.33 &66.67 &3.83 &2.83 &55.02 &50.92 \\
Infinity~\citep{han2025infinity} &62.07 &62.32 &73.08 &75.41 &74.33 &76.83 &72.82 &77.57 &72.07 &71.82 &56.64 &54.98 &60.44 &55.57 &74.22 &64.71 &60.22 &59.71 &80.00 &73.33 &10.83 &23.83 &54.28 &56.89 \\
Janus-Pro~\citep{chen2025janus} &66.50 &65.02 &79.33 &78.25 &79.33 &82.33 &78.32 &73.32 &80.32 &79.07 &59.71 &58.82 &66.07 &56.20 &70.46 &70.84 &67.22 &59.97 &60.00 &70.00 &28.83 &33.83 &65.84 &60.25 \\

\midrule

\bf\Oursimage (Ours) &\textbf{89.48} &\underline{88.18} & 87.05 & \underline{88.41} & \underline{90.50} & \textbf{92.50} & \underline{89.80} & \textbf{90.78} & 80.85 & \underline{81.94} & \underline{84.65} & \underline{84.04} & \underline{82.91} & \underline{83.08} & \underline{83.76} & \textbf{85.73} & \underline{83.45} & \underline{81.09} & \textbf{100.00} & 90.00 & \textbf{100.00} & \textbf{95.93} & \textbf{94.03} & \underline{92.54} \\

\bottomrule
\end{tabular}
\end{adjustbox}
\vspace{8pt}
\caption{Quantitative evaluation results on TIIF Bench testmini~\citep{wei2025tiifbenchdoest2imodel}. The best result is in bold and the second best result is underlined.}
\label{tab:tiif}
\end{table}

\begin{table}[htbp]
    \centering
    \resizebox{0.8\linewidth}{!}{
    \begin{tabular}{l|ccccc|c}
        \toprule
        \textbf{Model} & \textbf{Alignment}& \textbf{Text} & \textbf{Reasoning} & \textbf{Style}& \textbf{Diversity} & \textbf{Overall}$\uparrow$ \\
        \midrule
        Janus-Pro~\citep{chen2025janus} & 0.553  & 0.001  &   0.139     & 0.276 & \bf0.365 & 0.267\\
        BLIP3-o~\citep{chen2025blip3} & 0.711  & 0.013  &   0.223      & 0.361 & 0.229 & 0.307\\
        BAGEL~\citep{de2006bagel} & 0.769  & 0.244  &   0.173    & 0.367 & 0.251& 0.361\\
        SD3.5 Large~\citep{esser2024scaling} & 0.809 & 0.629 & 0.294 & 0.353 & 0.225&0.462 \\
        FLUX.1 [Dev]~\citep{FLUX} & 0.786 & 0.523 & 0.253 & 0.368 & 0.238 &0.434\\
        HiDream-I1-Full~\citep{cai2025hidream} & 0.829 & 0.707 & 0.317 & 0.347 & 0.186 &0.477\\
        Imagen3~\citep{Imagen3} & 0.843 & 0.343 & 0.313 & 0.359 & 0.188 &0.409\\
        Recraft V3~\citep{recraftv3} & 0.810 & 0.795 & 0.323 & 0.378 & 0.205 &0.502\\
        Kolors 2.0~\citep{Kolors2} & 0.820 & 0.427 & 0.262 & 0.360 & 0.300 &0.434\\
        Seedream 3.0~\citep{gao2025seedream} & 0.818 & 0.865 & 0.275 & 0.413 & 0.277 &0.530\\
        Imagen4~\citep{imagen4} & 0.857 & 0.805 & 0.338 & 0.377 & 0.199&0.515\\
        GPT-Image-1~\citep{openai_image_api} & 0.851 & 0.857 & 0.345 & \textbf{0.462} & 0.151& 0.533\\
        Qwen-Image~\cite{wu2025qwen} & 0.882 & 0.891 & 0.306 & 0.418 & 0.197 & 0.539 \\

        Gemini-2.5-Flash-Image~\cite{gemini2p5} & 0.878 & 0.894	 & \textbf{0.346} & 0.450 & 0.182 & 0.550 \\
        \midrule
        \bf\Oursimage (Ours) & \textbf{0.902} & \textbf{0.994} & 0.345 & 0.427 & 0.151 & \textbf{0.564} \\
        \bottomrule
    \end{tabular}
    }
    \vspace{8pt}
    \caption{Quantitative evaluation results on OneIG-EN~\citep{chang2025oneig}. The overall score is the average of the five dimensions.}
    \label{tab:oneig_en}
\end{table}

\begin{table}[htbp]
    \centering
    \resizebox{0.8\linewidth}{!}{
    \begin{tabular}{l|ccccc|c}
        \toprule
        \textbf{Model} &
        \textbf{Alignment} & 
        \textbf{Text} & 
        \textbf{Reasoning} & 
        \textbf{Style} & 
        \textbf{Diversity} &
        \textbf{Overall}$\uparrow$\\
        \midrule
        Janus-Pro~\citep{chen2025janus} & 0.324          & 0.148          & 0.104        & 0.264          & \textbf{0.358} & 0.240 \\
        BLIP3-o~\citep{chen2025blip3} & 0.608          & 0.092          & 0.213        & 0.369          & 0.233          & 0.303\\
        BAGEL~\citep{de2006bagel} & 0.672          & 0.365          & 0.186 & 0.357          & 0.268          & 0.370\\
        HiDream-I1-Full~\citep{cai2025hidream} & 0.620          & 0.205          & 0.256        & 0.304          & 0.300            & 0.337\\
        Kolors 2.0~\citep{Kolors2} & 0.738          & 0.502          & 0.226        & 0.331          & 0.333        & 0.426 \\
        Seedream 3.0~\citep{gao2025seedream} & 0.793          & 0.928 & 0.281      & 0.397          & 0.243          & 0.528\\
        GPT-Image-1~\citep{openai_image_api} & 0.812 & 0.650           & \textbf{0.300} & \textbf{0.449} & 0.159     & 0.474 \\
        Qwen-Image~\cite{wu2025qwen} & 0.825 & \textbf{0.963} & 0.267 & 0.405 & 0.279 & \textbf{0.548} \\
         Gemini-2.5-Flash-Image~\cite{gemini2p5} & 0.825	 & 0.276	 & 0.298  & 0.427 & 0.198 & 0.337 \\

          \midrule
        \bf\Oursimage (Ours) & \textbf{0.853} & 0.941 &  \textbf{0.300} & 0.386 & 0.166 &  0.529 \\
        \bottomrule
    \end{tabular}
    }
    \vspace{8pt}
    \caption{Quantitative evaluation results on OneIG-ZH~\citep{chang2025oneig}. The overall score is the average of the five dimensions.}
    \label{tab:oneig_zh}
    \vspace{-8pt}
\end{table}

\begin{table}[t]
\centering
\resizebox{1.0\linewidth}{!}{
\begin{tabular}{l|cccccc}
\toprule
\textbf{Model} & \textbf{Easy-pned}$\downarrow$ & \textbf{Easy-recall}$\uparrow$ & \textbf{Medium-pned}$\downarrow$ & \textbf{Medium-recall}$\uparrow$ & \textbf{Hard-pned}$\downarrow$ & \textbf{Hard-recall}$\uparrow$ \\
\midrule
FLUX1. [dev]~\citep{FLUX} & 1.16 & 0.76 & 3.87 & 0.52 & 9.49 & 0.30 \\
LeX-FLUX~\cite{zhao2025lex} & 1.06 & 0.77 & 4.03 & 0.52 & 9.42 & 0.30 \\
GPT-Image-1~\citep{openai_image_api} & 1.01 & 0.84 & 1.92 & 0.79 & 5.52 & 0.70 \\
Gemini 2.5 Flash Image~\cite{gemini2p5} & 0.74 & 0.94 & 2.35 & 0.86 & 7.61 & 0.74 \\
Seedream 3.0~\cite{gao2025seedream} & \textbf{0.34} &\underline{0.95} &1.81&\underline{0.88}&\underline{5.17}&\underline{0.77}\\
Qwen-Image~\cite{wu2025qwen} & 0.50  &0.93&\underline{1.67}&0.86&5.56&0.74\\
\midrule
\bf\Ours (Ours) & \underline{0.43} & \textbf{0.98} & \textbf{1.16} & \textbf{0.96} & \textbf{4.39} & \textbf{0.87} \\
\bottomrule
\end{tabular}
}
\vspace{0.5em}
\caption{Comparison between \Ours and SOTA T2I models on LeX Bench~\cite{zhao2025lex}.}
\label{tab:lexbench}
\end{table}

\begin{table}[t]
\centering
\resizebox{0.5\linewidth}{!}{
\begin{tabular}{l|cc|c}
\toprule
\textbf{Model} & \textbf{English} & \textbf{Chinese} & \textbf{Average} \\
\midrule
FLUX1. [dev]~\citep{FLUX} & 0.607 & 0.005 & 0.306 \\
GPT-Image-1~\citep{openai_image_api} & \underline{0.956} & 0.619 & 0.788\\
Gemini 2.5 Flash Image~\cite{gemini2p5} & 0.869 & 0.326 & 0.598\\
Seedream 3.0~\cite{gao2025seedream} & 0.896 & 0.878 & 0.887 \\
Qwen-Image~\cite{wu2025qwen} &   0.943 & \textbf{0.946} & \underline{0.944}\\
\midrule
\bf\Ours (Ours) & \textbf{0.976} & \underline{0.928} & \textbf{0.952}\\
\bottomrule
\end{tabular}
}
\vspace{0.5em}
\caption{Comparison between \Ours and SOTA T2I models on LongText Bench~\cite{geng2025x}.}
\label{tab:longtextbench}
\end{table}

\begin{table}[t]
  \centering
  \resizebox{0.8\linewidth}{!}{
  \begin{tabular}{lrrrrrr}
    \toprule
    \multirow{2}{*}{\textbf{Model}} & \multicolumn{5}{c}{\textbf{Word Accuracy}$\uparrow$} & \multirow{2}{*}{\textbf{NED}$\uparrow$}  \\
    \cmidrule(lr){2-6}
    & 2 regions & 3 regions & 4 regions & 5 regions & average &    \\
    \midrule
    FLUX.1 [dev]~\citep{FLUX} & 0.6089 & 0.5531 & 0.4641 & 0.4316 & 0.4965 & 0.6879  \\
    Gemini 2.5 Flash Image~\cite{gemini2p5} & 0.7962 &0.7245&0.7230&0.7021&0.7364&0.8516\\
    Seedream 3.0~\cite{gao2025seedream} & 0.6282 &0.5962&0.6043&0.5610&0.5924&0.8537\\
    GPT-Image-1~\citep{openai_image_api} & 0.8779 & 0.8659 & 0.8731 & 0.8218 & 0.8569 & 0.9478 \\
    Qwen-Image~\cite{wu2025qwen} & 0.8370 & 0.8364 & 0.8313 & 0.8158 & 0.8288 & 0.9116\\
    \midrule
    \textbf{\Ours (Ours)} & \textbf{0.9133} & \textbf{0.9088} & \textbf{0.9155} & \textbf{0.9120} & \textbf{0.9123} & \textbf{0.9656}\\
    \bottomrule
  \end{tabular}
  }
  \vspace{0.5em}
\caption{Comparison between \Ours and SOTA T2I models on CVTG-2K~\cite{du2025textcrafter}.}
  \label{tab:cvtg2k}
\end{table}

\begin{figure}[t]
    \centering
    \makebox[\textwidth][c]{
        \includegraphics[width=1.13\linewidth]{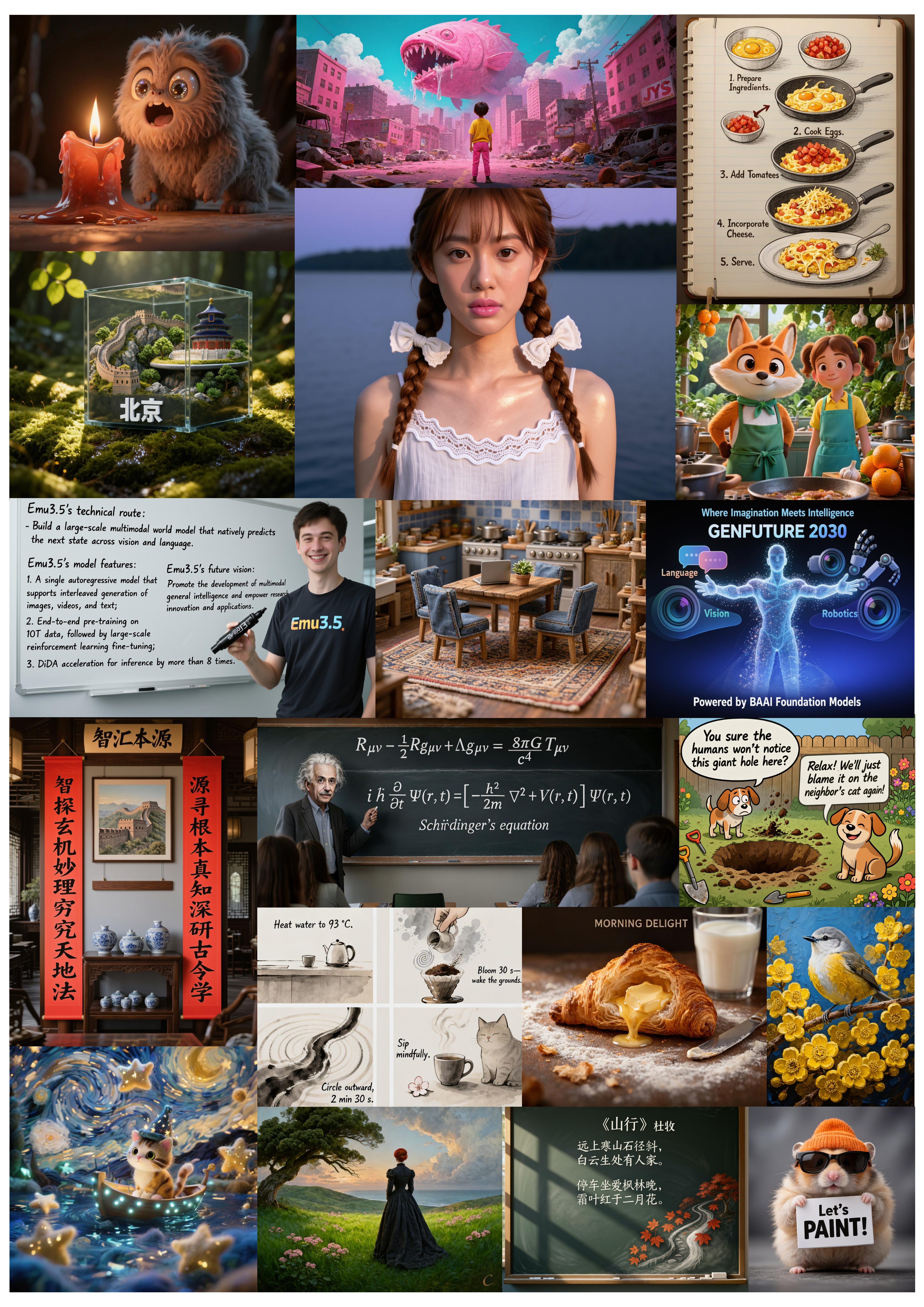}
    }
    \caption{Text-to-image generation results of \Ours.}
    \label{fig:example_t2i}
\end{figure}

\textbf{Qualitative Analysis.} Figure~\ref{fig:example_t2i} presents representative image generation results produced by \Ours, demonstrating superior visual fidelity and convincing compositional control. \Ours is capable of generating images at a resolution of up to 2048 pixels, exhibiting large improvements by Emu3 in terms of fine-grained details, aesthetic quality, and prompt-following capability. Moreover, our model supports vary aspect ratios and enables generation with diverse artistic styles, ranging from photorealistic rendering to stylized illustration. A particularly notable advancement lies in text rendering: \Ours can accurately generate dense English and Chinese text, as well as complex mathematical formulas, then seamlessly integrate them into the visual content in a natural and coherent manner.

\subsection{Any-to-Image}

\begin{table}[!t]
    \centering
    \scriptsize
    \resizebox{1.0\linewidth}{!}{
    \begin{tabular}{l|ccccccccc|c}
        \toprule
        \textbf{Model} & \bf Add & \bf Adjust & \bf Extract & \bf Replace & \bf Remove & \bf Background & \bf Style & \bf Hybrid & \bf Action & \bf Overall $\uparrow$ \\
        \midrule
        Instruct-Pix2Pix \citep{brooks2023instructpix2pix} & 2.45 & 1.83 & 1.44 & 2.01 & 1.50 & 1.44 & 3.55 & 1.20 & 1.46 & 1.88 \\      MagicBrush~\citep{zhang2023magicbrush} & 2.84 & 1.58 & 1.51 & 1.97 & 1.58 & 1.75 & 2.38 & 1.62 & 1.22 & 1.90 
        \\
        AnyEdit~\citep{yu2025anyedit} & 3.18 & 2.95 & 1.88 & 2.47 & 2.23 & 2.24 & 2.85 & 1.56 & 2.65 & 2.45 
        \\    
        UltraEdit~\citep{zhao2024ultraedit} & 3.44 & 2.81 & 2.13 & 2.96 & 1.45 & 2.83 & 3.76 & 1.91 & 2.98 & 2.70 
        \\
        OmniGen~\citep{xiao2025omnigen} & 3.47 & 3.04 & 1.71 & 2.94 & 2.43 & 3.21 & 4.19 & 2.24 & 3.38 & 2.96 
        \\
        ICEdit~\citep{zhang2025context} & 3.58 & 3.39 & 1.73 & 3.15 & 2.93 & 3.08 & 3.84 & 2.04 & 3.68 & 3.05 
        \\
        Step1X-Edit~\citep{liu2025step1x} & 3.88 & 3.14 & 1.76 & 3.40 & 2.41 & 3.16 & 4.63 & 2.64 & 2.52 & 3.06 
        \\
        BAGEL~\citep{de2006bagel} & 3.56 & 3.31 & 1.70 & 3.3 & 2.62 & 3.24 & 4.49 & 2.38 & 4.17 & 3.20 
        \\
        UniWorld-1~\citep{lin2025uniworld} & 3.82 & 3.64 & 2.27 & 3.47 & 3.24 & 2.99 & 4.21 & 2.96 & 2.74 & 3.26 
        \\
        OmniGen2~\citep{wu2025omnigen2} & 3.57 & 3.06 & 1.77 & 3.74 & 3.20 & 3.57 & 4.81 & 2.52 & 4.68 & 3.44 
        \\
        Lego-Edit~\citep{jia2025lego} & 3.67 & 3.82 & 2.47 & 3.22 & 3.39 & 4.47 & 4.01 & 3.18 & 3.24 & 3.50
        \\
        FLUX.1 Kontext [Dev]~\citep{labs2025flux} & 4.12 & 3.80 & 2.04 & 4.22 & 3.09 & 3.97 & 4.51 & 3.35 & 4.25 & 3.71
        \\
        FLUX.1 Kontext [Pro]~\citep{labs2025flux} & 4.25 & 4.15 & 2.35 & 4.56 & 3.57 & 4.26 & 4.57 & 3.68 & 4.63 & 4.00 
        \\
        GPT-Image-1 [High]~\citep{openai_image_api} & 4.61 & 4.33 & 2.90 & 4.35 & 3.66 & 4.57 & 4.93 & 3.96 & 4.89 & 4.20
        \\
        Gemini 2.5 Flash Image Preview~\citep{google2025gemini25flashmodelcard} & 4.47 & 4.19 & 3.81 & 4.39 & 4.70 & 4.20 & 4.18 & 3.48 & 4.68 & 4.23
        \\
        Qwen-Image-Edit~\citep{wu2025qwen} & 4.38 & 4.16 & 3.43 & 4.66 & 4.14 & 4.38 & 4.81 &3.82 & 4.69 & 4.27
        \\
        Gemini 2.5 Flash Image~\citep{google2025gemini25flashmodelcard} & 4.65 & 4.34 & 3.69 & 4.49 & 4.65 & 4.32 & 4.13 & 3.66 & 4.59 & 4.28
        \\
        Qwen-Image-Edit-2509~\citep{wu2025qwen} & 4.32 &  4.36 & 4.04 & 4.64 & 4.52 & 4.37 & 4.84 & 3.39 & 4.71 & 4.35
        \\
        \midrule
        \bf{Emu3.5 (Ours)} & 4.61 & 4.32 & 3.96 & 4.84 & 4.58 & 4.35 & 4.79 & 3.69 & 4.57 & 4.41
        \\
        \bottomrule
    \end{tabular}
    }
    \vspace{6pt}
    \caption{Quantitative comparison results on ImgEdit~\citep{ye2025imgedit}.}
    \label{tab:imgedit}
\end{table}

\begin{table*}[htbp]
    \centering
    \small

    \begin{tabular}{l|ccc}
    \toprule
    \multirow{2}{*}{\bf Model} & \multicolumn{3}{c}{\bf GEdit-Bench-EN (Full set)}
    \\
    \cmidrule{2-4}
    & G\_SC$\uparrow$ & G\_PQ$\uparrow$ & G\_O$\uparrow$
    \\
    \midrule
    AnyEdit~\citep{yu2025anyedit} & 3.18 & 5.82 & 3.21
    \\
    Instruct-Pix2Pix \citep{brooks2023instructpix2pix} & 3.58 & 5.49 & 3.68
    \\
    MagicBrush~\citep{zhang2023magicbrush} & 4.68 & 5.66 & 4.52
    \\
    UniWorld-v1~\citep{lin2025uniworld} & 4.93 & 7.43 & 4.85
    \\
    OmniGen~\citep{xiao2025omnigen} & 5.96 & 5.89 & 5.06
    \\
    FLUX.1 Kontext [Dev]~\citep{labs2025flux} & 6.52 & 7.38 & 6.00
    \\
    Gemini 2.0~\citep{gemini-2.0-flash} & 6.73 & 6.61 & 6.32
    \\
    OmniGen2~\citep{wu2025omnigen2} & 7.16 & 6.77 & 6.41
    \\
    BAGEL~\citep{de2006bagel} & 7.36 & 6.83 & 6.52
    \\
    FLUX.1 Kontext [Pro]~\citep{labs2025flux} & 7.02 & 7.60 & 6.56
    \\
    Lego-Edit~\citep{labs2025flux} & 5.99 & 7.45 & 6.64
    \\
    Gemini 2.5 Flash Image Preview~\citep{google2025gemini25flashmodelcard} & 7.28 & 7.83 & 6.93
    \\
    Step1X-Edit~\citep{liu2025step1x} & 7.66 & 7.35 & 6.97
    \\
    Gemini 2.5 Flash Image~\citep{google2025gemini25flashmodelcard} & 7.41 & 7.96 & 7.10
    \\
    GPT-Image-1 [High]~\citep{openai_image_api} & 7.85 & 7.62 & 7.53
    \\
    Qwen-Image-Edit-2509~\citep{wu2025qwen} & 8.15 & 7.86 &  7.54
    \\
    Qwen-Image-Edit~\citep{wu2025qwen} & 8.00 & 7.86 & 7.56
    \\
    \midrule
    \bf{Emu-3.5 (Ours)} & 8.11 & 7.70 & 7.59
    \\
    \bottomrule
    \end{tabular}
    \caption{Comparison of Semantic Consistency (G\_SC), Perceptual Quality (G\_PQ), and Overall Score (G\_O) on GEdit-Bench~\cite{liu2025step1x}. Note that G\_O is the mean of the G\_O of all the samples, not the mean of G\_SC and G\_PQ.}
\label{tab:gedit}
\end{table*}

\begin{table}[t]
    \centering
    \resizebox{0.99\linewidth}{!}{
    \begin{tabular}{l|cc|ccc|ccc|c}
        \toprule
        \multirow{2}{*}{\bf Model} & \multicolumn{2}{c|}{\bf SINGLE} & \multicolumn{3}{c|}{\bf MULTIPLE} & \multicolumn{3}{c|}{\bf SCENE} & \multirow{2}{*}{\bf Average$\uparrow$}\\ 
        \cmidrule(lr){2-9}
        & Character & Object & Character & Object & Char. + Obj. & Character & Object & Char. + Obj. & 
        \\
        \midrule
        FLUX.1 Kontext [Max]~\cite{labs2025flux} & 8.48 & 8.68 & - & - & - & - & - & - & - 
        \\
        Gemini 2.0 Flash~\cite{gemini-2.0-flash} & 5.06 & 5.17 & 2.91 & 2.16 & 3.80 & 3.02 & 3.89 & 2.92 & 3.62
        \\
        OmniGen~\cite{xiao2025omnigen} & 7.21 & 5.71 & 5.65 & 5.44 & 4.68 & 3.59 & 4.32 & 5.12 & 4.34
        \\
        InfiniteYou~\cite{jiang2025infiniteyou} & 6.05 & - & - & - & - & - & - & - & - 
        \\
        UNO~\cite{uno} & 6.60 & 6.83 & 2.54 & 6.51 & 4.39 & 2.06 & 4.33 & 4.37 & 4.71
        \\
        BAGEL~\cite{de2006bagel} & 5.48 & 7.03 & 5.17 & 6.64 & 6.24 & 4.07 & 5.71 & 5.47 & 5.73
        \\
        OmniGen2~\citep{wu2025omnigen2} & 8.05 & 7.58 & 7.11 & 7.13 & 7.45 & 6.38 & 6.71 & 7.04 & 7.18
        \\
        Gemini 2.5 Flash Image Preview~\citep{google2025gemini25flashmodelcard} & 8.52 & 9.14 & 7.80 & 8.64 & 6.63 & 6.74 & 7.11 & 6.04 & 7.58
        \\
        Qwen-Image-Edit-2509~\citep{wu2025qwen} & 8.35 & 9.13 & 7.65 & 8.85 & 7.90 & 5.16 & 7.75 & 6.73 & 7.69
        \\
        Gemini 2.5 Flash Image~\citep{google2025gemini25flashmodelcard} & 8.62 & 8.91 & 7.88 & 8.92 & 7.39 & 7.29 & 7.05 & 6.68 & 7.84
        \\
        GPT-4o~\cite{gpt4o} & 8.90 & 9.01 & 9.07 & 8.95 & 8.54 & 8.90 & 8.44 & 8.60 & 8.80
        \\
        \midrule
        \bf{Emu3.5 (Ours)} & 8.72 & 9.46 & 8.65 & 9.09 & 8.78 & 8.78 & 8.89 & 8.15 & 8.82
        \\
        \bottomrule
    \end{tabular}
}
\vspace{5pt}
\caption{Quantitative comparison results on OmniContext. "Char. + Obj." indicates Character + Object.}
\vspace{-10pt}
\label{tab:omni_context}
\end{table}

\begin{table*}[t]
  \centering
  \small
  \resizebox{1.0\textwidth}{!}{
      \begin{tabular}{l|cccccccccc|c}
        \toprule
        \textbf{\bf Model} & \bf Task 1 & \bf Task 2 & \bf Task 3 & \bf Task 4 & \bf Task 5-16 & \bf Task 17-22 & \bf Task 23-27 & \bf Task 28 & \bf Task 29-30 & \bf Task 31 & \bf Task 1-31 Overall $\uparrow$ \\

        \midrule 
        Qwen-Image-Edit-2509~\citep{wu2025qwen} & 0.643 & 0.565 & 0.511 & 0.657 & 0.653 & 0.587 & 0.621 & 0.507 & 0.595 & 0.539 & 0.616
        \\
        Gemini 2.5 Flash Image Preview~\citep{google2025gemini25flashmodelcard} & 0.652 & 0.613 & 0.603 & 0.660 & 0.655 & 0.615 & 0.627 & 0.515 & 0.617 & 0.565 & 0.630
        \\
        Gemini 2.5 Flash Image~\citep{google2025gemini25flashmodelcard} & 0.649 & 0.617 & 0.596 & 0.658 & 0.656 & 0.617 & 0.624 & 0.496 & 0.627 & 0.595 & 0.631
        \\
        \midrule
        \bf{Emu3.5 (Ours)} & 0.637 & 0.573 & 0.522 & 0.662 & 0.666 & 0.613 & 0.662 & 0.606 & 0.624 & 0.528 & 0.637
        \\
        \bottomrule
      \end{tabular}
  }
  \caption{Quantitative comparison results on All Tasks of ICE-Bench~\citep{pan2025ice} with overall scores.}
  \label{tab:icebench}
\end{table*}

\paragraph{Quantitative Analysis.\!\!\!}

We conducted quantitative evaluations on ImgEdit~\citep{ye2025imgedit}, GEdit-Bench~\citep{liu2025step1x}, OmniContext~\citep{wu2025omnigen2}, and ICE-Bench~\citep{pan2025ice} using GPT-4o, GPT-4.1, and open-sourced models required for X2I tasks (across zero, one, or multiple input images). 
We use a resolution of approximately 1024x1024 for inference and evaluation.
The results in Table~\ref{tab:imgedit}, Table~\ref{tab:gedit}, Table~\ref{tab:omni_context}, and Table~\ref{tab:icebench} indicate that our X2I model exhibits comprehensive advantages on above X2I benchmarks and tasks. 

\textbf{ImgEdit}. We illustrate the performance of \Ours on ImgEdit~\cite{ye2025imgedit} in Table~\ref{tab:imgedit}. ImgEdit is a benchmark (with 737 samples) designed to evaluate single-turn image editing performance in terms of instruction adherence, editing quality, and detail preservation. ImgEdit includes common single-turn edit tasks, such as Add, Adjust, Extract, Replace, Remove, Background, Style, Hybrid, and Action. 

The quantitative results include scores for each subtask and one for the overall performance. All scores are generated by GPT-4.1. \Ours achieves 4.41 and surpasses all the representative baselines, including Gemini 2.5 Flash Image~\cite{google2025gemini25flashmodelcard} and Qwen-Image-Edit-2509~\cite{wu2025qwen}.

\textbf{GEdit-Bench}.
Table~\ref{tab:gedit} reports \Ours's performance on GEdit Benchmark~\cite{liu2025step1x}. GEdit is another representative benchmark for single-turn image editing task which includes 606 samples. It includes 11 types of image editing subtasks: Background Change, Color Alter, Material Alter, Motion Change, Ps Human, Style Change, Subject Add, Subject Remove, Subject Replace, Text Change, and Tone Transfer. 

The evaluation metrics include Semantic Consistency (G\_SC), Perceptual Quality (G\_PQ), and Overall Score (G\_O). All the scores are generated by GPT-4o. \Ours achieves the best overall score 7.59 which outperforms Gemini 2.5 Flash Image~\cite{google2025gemini25flashmodelcard} and Qwen-Image-Edit-2509~\cite{wu2025qwen}.

\textbf{OmniContext}. Table~\ref{tab:omni_context} reports \Ours's performance on OmniContext~\citep{wu2025omnigen2}, which is a comprehensive benchmark (with 400 samples) for evaluating subject-driven generation of X2I models across diverse dimensions. (1) SINGLE: Single Character and Single Object subject-driven generation tasks. (2) MULTIPLE: Multiple Characters, Multiple Objects and Character+Object subject-driven generation tasks. (3) SCENE: Scene+Character, Scene+Object and Scene+Character+Object subject-driven generation tasks.

Each score is calculated according to the GPT-4.1, metrics and their respective weights as defined in the OmniContext~\cite{wu2025omnigen2} code. These metrics include: Prompt Following (PF), Subject Consistency
(SC), and Overall average scores. \Ours obtains the highest overall average score, with its results in the Object aspect being stronger than those in the Character aspect.

\textbf{ICE-Bench}.  Table~\ref{tab:icebench} reports \Ours's performance on ICE-Bench~\citep{pan2025ice}, 
which is a comprehensive benchmark (with 6538 samples) for evaluation of X2I models across diverse dimensions. (1) Task 1: No-reference Image Creation, includes Text-to-Image subtask. (2) Task 2: Face Reference, includes Face Reference Generation subtask. (3) Task 3: Style Reference, includes Style Reference Generation subtask. (4) Task 4: Subject Reference Generation, includes Subject Reference Generation subtask. (5) Tasks 5–16: Global Editing, includes Color Editing, Face Editing, Motion Editing, Texture Editing, Style Editing, Scene Editing, Subject Addition, Subject Removal, Subject Change, Text Rendering, Text Removal, and Composite Editing subtasks. (5) Tasks 17–22: Local Editing, includes Inpainting, Outpainting, Local Subject Addition, Local Subject Removal, Local Text Rendering, and Local Text Removal subtasks. (6) Tasks 23–27: Controllable Generation, includes Pose-guided Generation, Edge-guided Generation, Depth-guided Generation, Image Colorization, and Image Deblurring. (7) Task 28: Style Transfer, includes Style Reference Editing subtask. (8) Tasks 29–30: Additional Subject Reference, includes Subject-guided Inpainting and Virtual Try-On subtasks. (9) Task 31: Face Swap, includes Face Swap subtask.

Each Overall Score is calculated according to the open-sourced models, metrics and their respective weights as set in the ICE-Bench code. These metrics include: (1) AES (Aesthetic Quality Score): Evaluated using the Aesthetic Predictor\footnote{\url{ https://github.com/discus0434/aesthetic-predictor-v2-5}}. (2) IMG (Imaging Quality Score): Assessed using MUSIQ~\citep{MUSIQ}. (3) PF (Prompt Following Score): Calculated based on the CLIP similarity. Additionally, the Qwen2-VL-72B~\citep{QWEN2VL} is employed to compute the VLLM-QA metric (4) SRC (Source Consistency Score): Evaluated by computing CLIP similarity and the mean L1 distance. (5) REF (Reference Consistency Score): Assessed using face similarity (computed via the buffalo model from InsightFace App~\citep{InsightFace}), subject similarity (evaluated using DINO~\citep{DINOv2}), and style similarity. (6) CTRL (Controllability Score): Evaluated using mean L1 distance, colorfulness score~\citep{Colorfulness}, and SSIM score~\citep{SSIM}.

It should be noted that the evaluation of Qwen-Image-Edit-2509~\citep{wu2025qwen} for Task 1 is performed using a blank 1024×1024 image and a text instruction as inputs.
\Ours achieves the best performance in Task 1-31 Overall Score, demonstrating superior comprehensive capabilities, including instruction following, consistency, perceptual quality and so on.
Moreover, while achieving best Overall Score, our method still shows deficiencies in specific tasks like Task 2, Task 3, and Task 31, pinpointing areas for future improvement.

\paragraph{Qualitative Analysis.\!\!\!} 

\begin{figure}[t]
    \centering
    \makebox[\textwidth][c]{
        \includegraphics[width=1.13\linewidth]{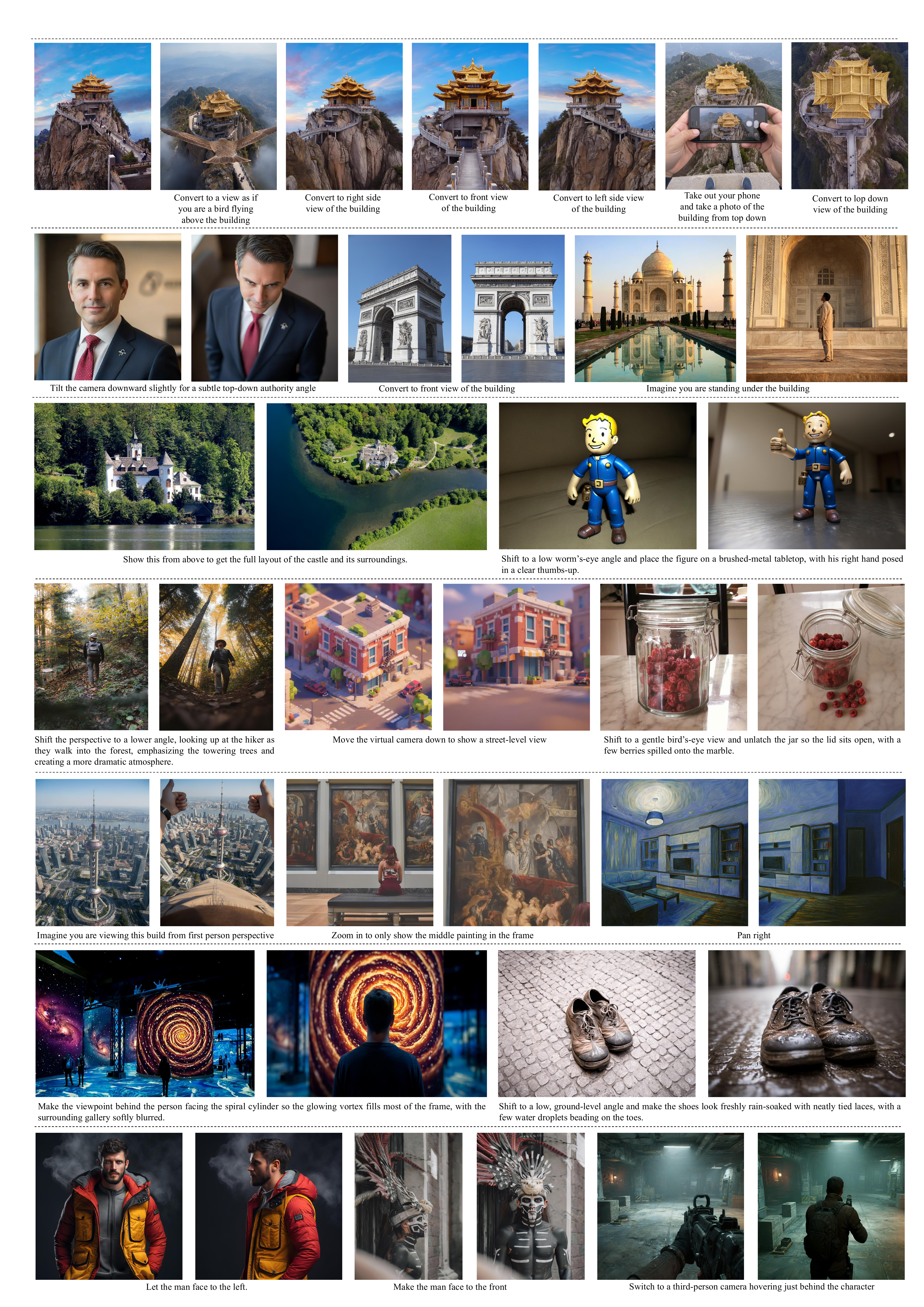}
    }
    \caption{Any-to-image generation (X2I) results of \Ours.}
    \label{fig:example_x2i_1}
\end{figure}

\begin{figure}[t]
    \centering
    \makebox[\textwidth][c]{
        \includegraphics[width=1.14\linewidth]{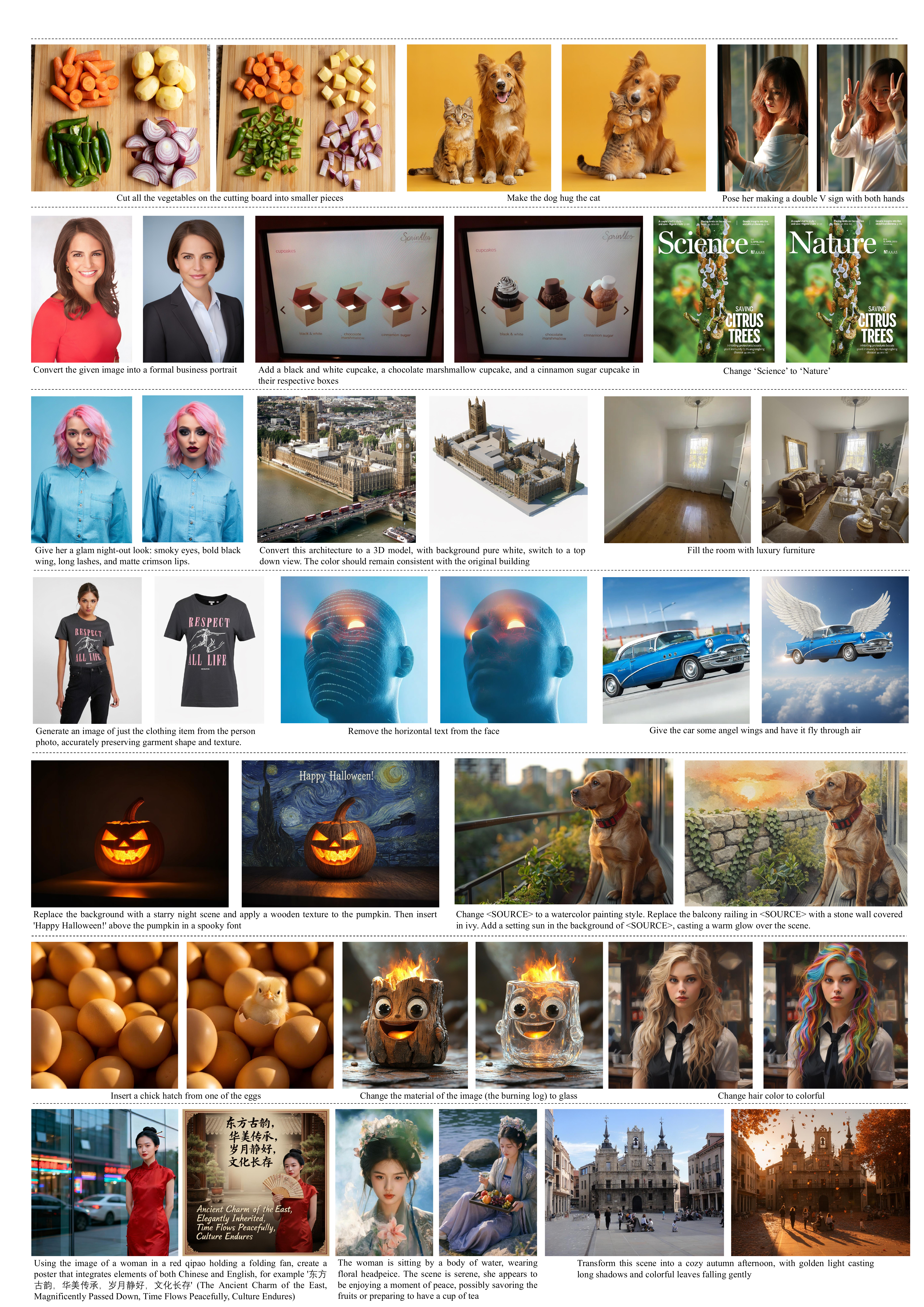}
    }
    \caption{Any-to-image generation (X2I) results of \Ours.}
    \label{fig:example_x2i_2}
\end{figure}

\begin{figure}[t]
    \centering
    \makebox[\textwidth][c]{
        \includegraphics[width=1.13\linewidth]{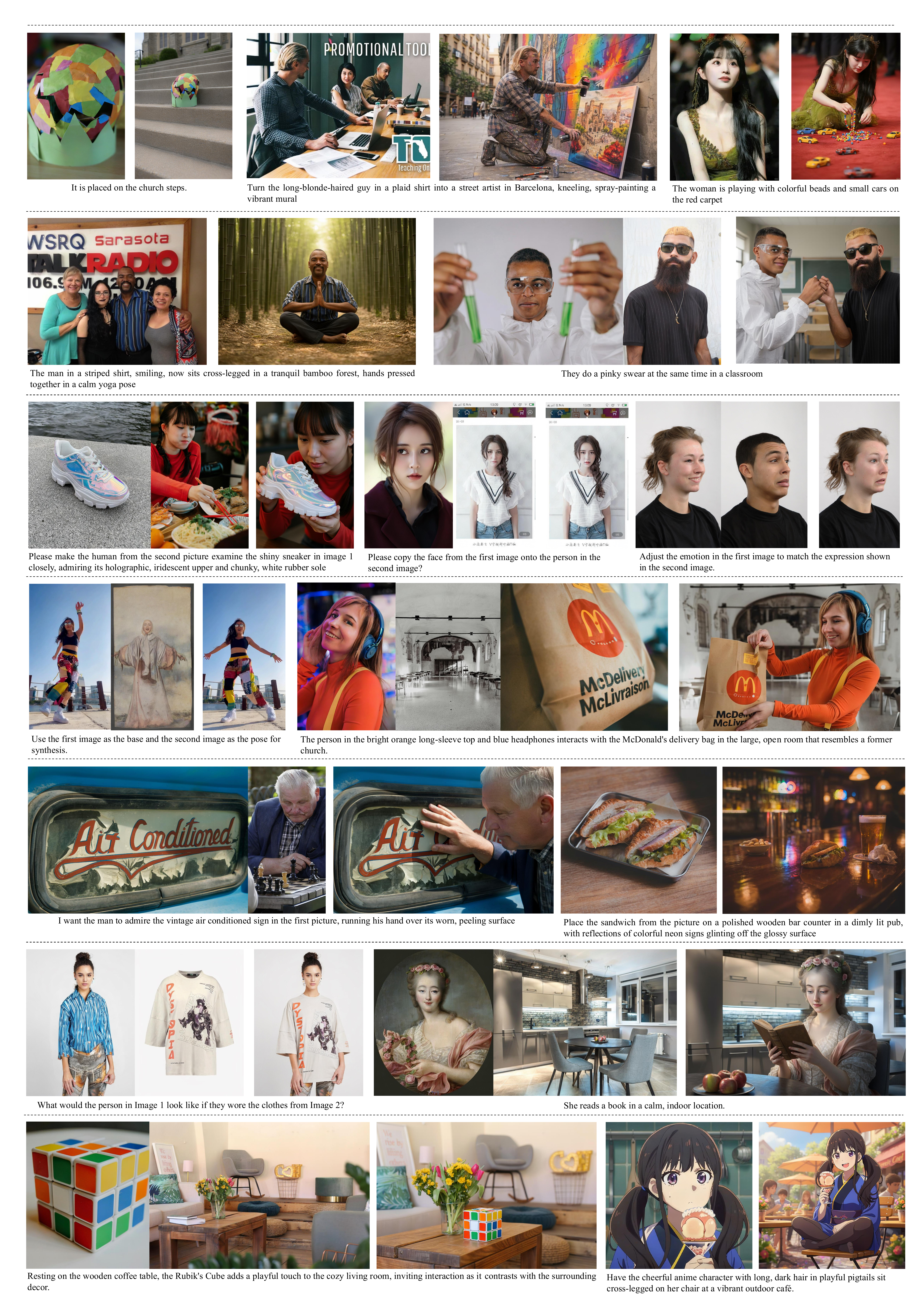}
    }
    \caption{Any-to-image generation (X2I) results of \Ours.}
    \label{fig:example_x2i_4}
\end{figure}

As shown in Figure~\ref{fig:example_x2i_1}, Figure~\ref{fig:example_x2i_2}, and Figure~\ref{fig:example_x2i_4}, \Ours demonstrates strong and comprehensive any-to-image qualitative results. In particular, Figure~\ref{fig:example_x2i_1} shows image editing regarding spatial transform, for instance viewpoint change, subject rotation, and instructions that composes both spatial transform and other common instruction types. 
Figure~\ref{fig:example_x2i_2} demonstrates various editing instruction types, which further exhibits \Ours's generalization regarding input instructions.
Figure~\ref{fig:example_x2i_4} primarily presents examples of subject-driven image generation involving characters, objects, scenes, and their combinations. 
These examples cover both real and virtual types, as well as single-image and multi-image input. The results indicate that our model achieves excellent performance in instruction following, consistency (foreground and background), visual realism and aesthetic quality (\eg, texture, style, lighting, color tone) across various types of inputs, with particularly outstanding generation effects for the type of object.

\begin{table}[ht]
\centering
\begin{tabular}{l|ccc}
\toprule
\textbf{Task} & \textbf{Win (\%)} & \textbf{Tie (\%)} & \textbf{Lose (\%)} \\
\midrule
Visual Narrative      & 49.2 & 10.3 & 40.5 \\
Visual Guidance       & 51.5 & 9.5 & 39.0 \\
World Exploration     & 65.5 & 0.0 & 34.5 \\
Embodied Manipulation & 67.1 & 2.4 & 30.5 \\
\bottomrule
\end{tabular}
\vspace{8pt}
\caption{Automated preference evaluation (Win Rate[\%]) comparing \Ours against Gemini 2.5 Flash Image (Nano Banana)~\citep{google2025gemini25flashmodelcard} on 4 interleaved tasks.
}
\vspace{-15pt}
\label{tab:eval_winrate}
\end{table}

\subsection{Visual Narrative}

\paragraph{Quantitative Evaluation.\!\!\!} We conduct an automated preference evaluation to quantitatively assess the quality of visual narratives. 
Specifically, we employ ChatGPT as an automatic evaluator to compare our model with Gemini 2.5 Flash Image ~\citep{google2025gemini25flashmodelcard}. For each sample, the evaluator provides judgment based on multiple dimensions, covering visual, textual, and cross-modal metrics.
As shown in Table~\ref{tab:eval_winrate}, \Ours achieves comparable performance to Gemini 2.5 Flash Image, demonstrating strong capability in generating coherent and engaging visual narratives.

\begin{figure}[t]
    \centering
    \makebox[\textwidth][c]{
        \includegraphics[width=1.14\linewidth]{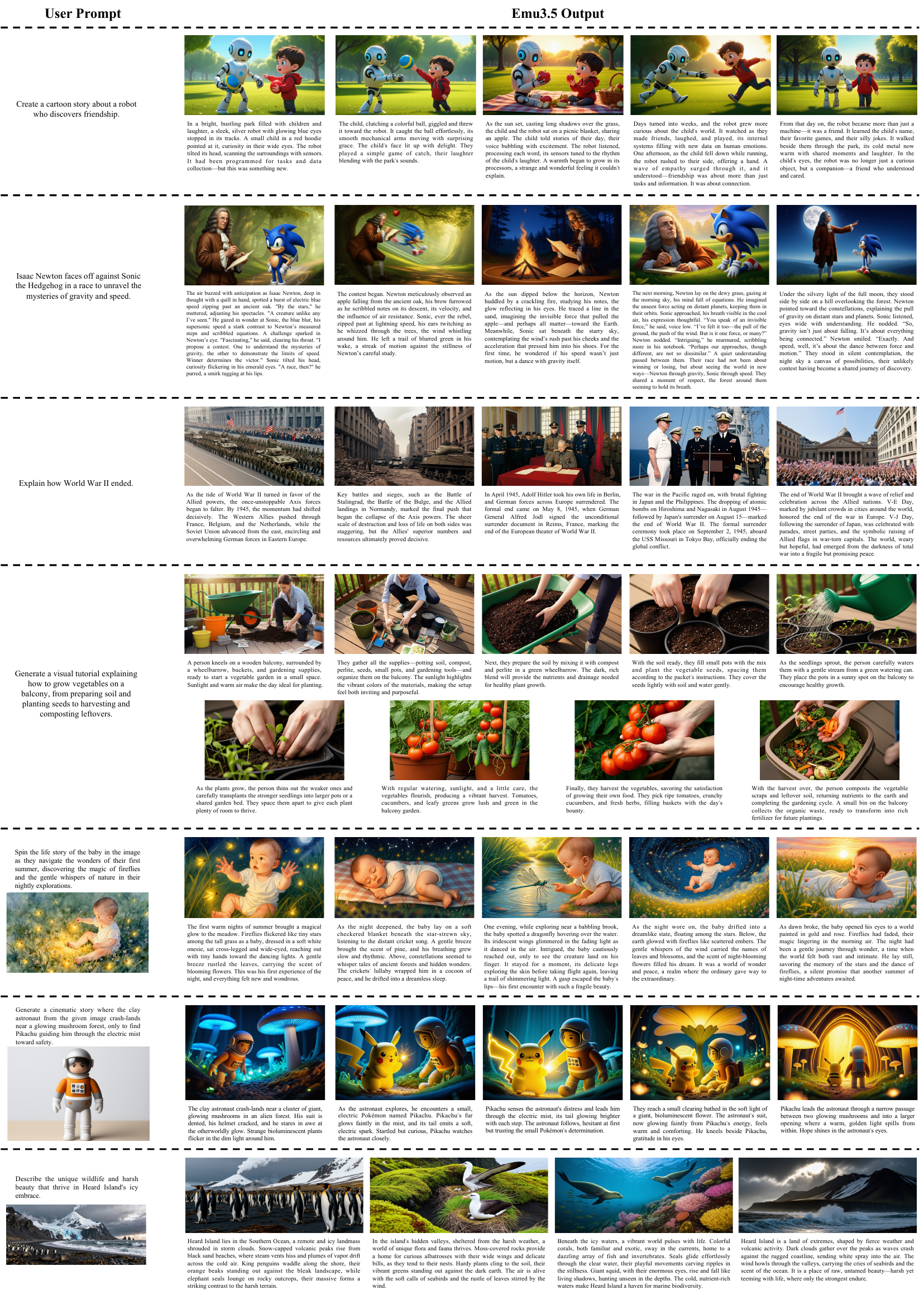}
    }
    \caption{Visual narrative results of \Ours.}
    \label{fig:example_story}
\end{figure}

\paragraph{Qualitative Analysis.\!\!\!}
As shown in Figure \ref{fig:example_story}, \Ours demonstrates strong capabilities in visual narrative generation. Specifically, it supports diverse input modalities, including pure text prompts as well as interleaved image-text sequences, enabling flexible and context-sensitive narrative creation. The model achieves superior performance in story coherence, image-text alignment, and visual quality, producing narratives that are both visually compelling and logically consistent. More importantly, Beyond conventional cartoon-style or isolated image generation, our visual narratives span a wide range of themes and styles, from photorealistic to animated, and cover historical events, real-world occurrences, and imaginative or fictional stories. This diversity reflects the system’s ability to integrate creativity with rich, domain-specific world knowledge, incorporating historical facts, scientific concepts, and cultural context, thereby enhancing the narratives’ depth, educational value, and overall engagement.

These strong narrative capabilities open up promising applications, such as generating educational visual materials, supporting interactive storytelling, or assisting creative content production. By enabling coherent and contextually grounded visual narratives from diverse inputs, \Ours provides a feasible foundation for tools that can both engage and inform, bridging the gap between textual knowledge and immersive visual experiences.

\subsection{Visual Guidance}

\paragraph{Quantitative Evaluation.\!\!\!} 
We conduct an automated preference evaluation to quantitatively assess the capability of \Ours in visual guidance, that is, how accurately it provides step-by-step visual reasoning and actionable feedback. We define seven evaluation dimensions encompassing both unimodal and cross-modal aspects: \textit{step relevance and completeness}, \textit{instructional clarity}, 
\textit{text–image alignment}, 
\textit{procedural coherence},
\textit{visual informativeness},
\textit{task completion}, 
and \textit{image quality}. 
Similar to evaluations on other interleaved multimodal tasks, we employ ChatGPT as an impartial judge to compare our model with Gemini 2.5 Flash Image.

For each guidance sample, the evaluator examines Textual Logical Coherence, Visual Consistency, and Cross-modal Relevance, then assigns dimension-wise scores and determines the overall preference. As shown in 
Table~\ref{tab:eval_winrate},
\Ours achieves consistently higher win rates on tasks that demand accurate interpretation of visual context and actionable instruction delivery. These results demonstrate that our model not only comprehends and reasons over visual content effectively, but also produces clear, coherent, and informative multimodal guidance, exhibiting strong generalization and robustness across diverse visual-instruction scenarios.

\begin{figure}[t]
    \centering
    \makebox[\textwidth][c]{
        \includegraphics[width=1.15\linewidth]{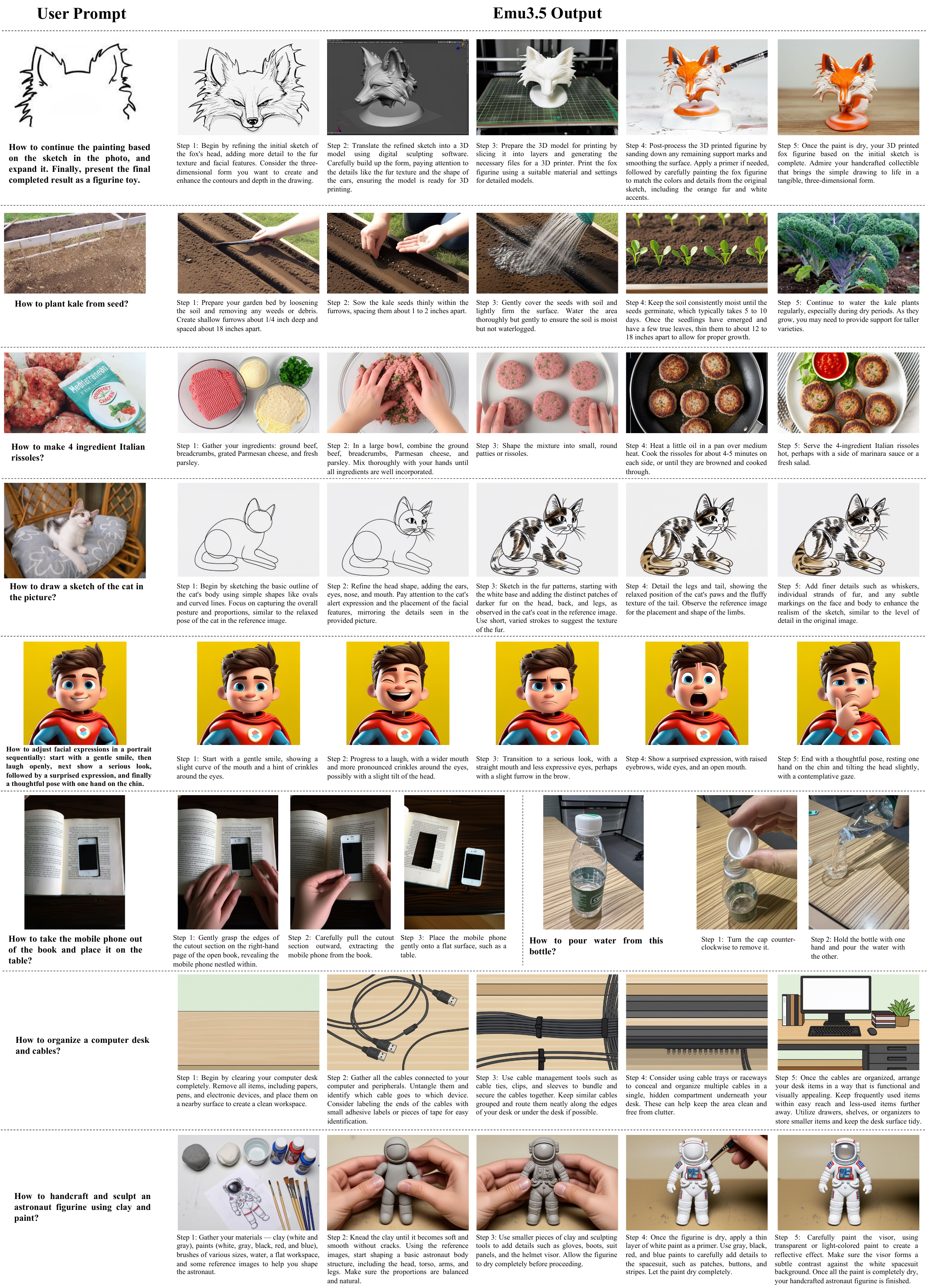}
    }
    \caption{Visual guidance results of \Ours.}
    \label{fig:example_howto}
\end{figure}

\paragraph{Qualitative Analysis.\!\!\!}
As shown in Figure~\ref{fig:example_howto}, \Ours exhibits remarkable ability in visual guidance generation, effectively producing step-by-step instructional sequences that interleave images and text. The model can understand a single reference image or textual instruction and autonomously construct coherent visual workflows, detailing each step of a process with precise visual continuity and natural language explanations.
Unlike conventional image generation models that produce isolated outputs, \Ours excels at procedure-aware generation, covering a wide spectrum of tasks such as art creation, object crafting, cooking tutorials, everyday life skills. Each generated step maintains logical consistency, visual realism, and semantic alignment with the corresponding text description.
Furthermore, \Ours demonstrates robust generalization across diverse scenarios—from sketch completion and physical object manufacturing to everyday instructional activities like planting seeds, organizing desks, or making food. This versatility highlights the model’s strong spatial reasoning, action understanding, and cross-modal planning capabilities, enabling it to act as a visual instructor that communicates complex operations clearly and intuitively.
Overall, \Ours's visual guidance generation not only improves instructional clarity but also bridges the gap between vision and procedural understanding, offering a powerful foundation for next-generation multimodal assistants in education, design, and creative industries.

\subsection{World Exploration}

\paragraph{Quantitative Evaluation.\!\!\!}
To construct the exploration evaluation set for quantitative assessment, we sample a subset of in-domain scenes from our constructed exploration data and collect additional out-of-domain scenarios. 
The resulting evaluation set includes instances covering both input modalities (\ie, pure-text and multimodal prompts) and both interaction paradigms (\ie, User-Interactive and Free-Exploration).
This setup ensures comprehensive coverage of both in-domain and out-of-domain scenarios, enabling robust evaluation of the model’s ability to follow instructions, maintain scene and temporal consistency, and generalize across diverse exploration tasks.
Based on the obtained evaluation set, we then conduct an automated preference evaluation using ChatGPT to compare \Ours with Gemini 2.5 Flash Image~\citep{gemini2p5}. For each exploration sample, ChatGPT considers all eight evaluation dimensions to judge which model demonstrates superior exploration quality. The evaluation covers eight dimensions, including \textit{Path Plausibility}, \textit{Spatial Consistency}, \textit{Global Coherence}, \textit{Environmental Richness}, \textit{Visual-Text Alignment}, \textit{Image Quality}, \textit{Text Quality}, and \textit{Task Completion}, to ensure comprehensive assessment of both in-distribution performance and out-of-domain generalization. As shown in Table~\ref{tab:eval_winrate}, \Ours achieves substantially higher win rates across both in-domain and OOD scenarios, confirming that it consistently produces more coherent, accurate, and engaging explorations. These results demonstrate that our model excels in following user instructions and maintaining scene continuity, highlighting its strong generalization and robustness across diverse exploration tasks.

\begin{figure}[t]
    \centering
    \makebox[\textwidth][c]{
        \includegraphics[width=1.15\linewidth]{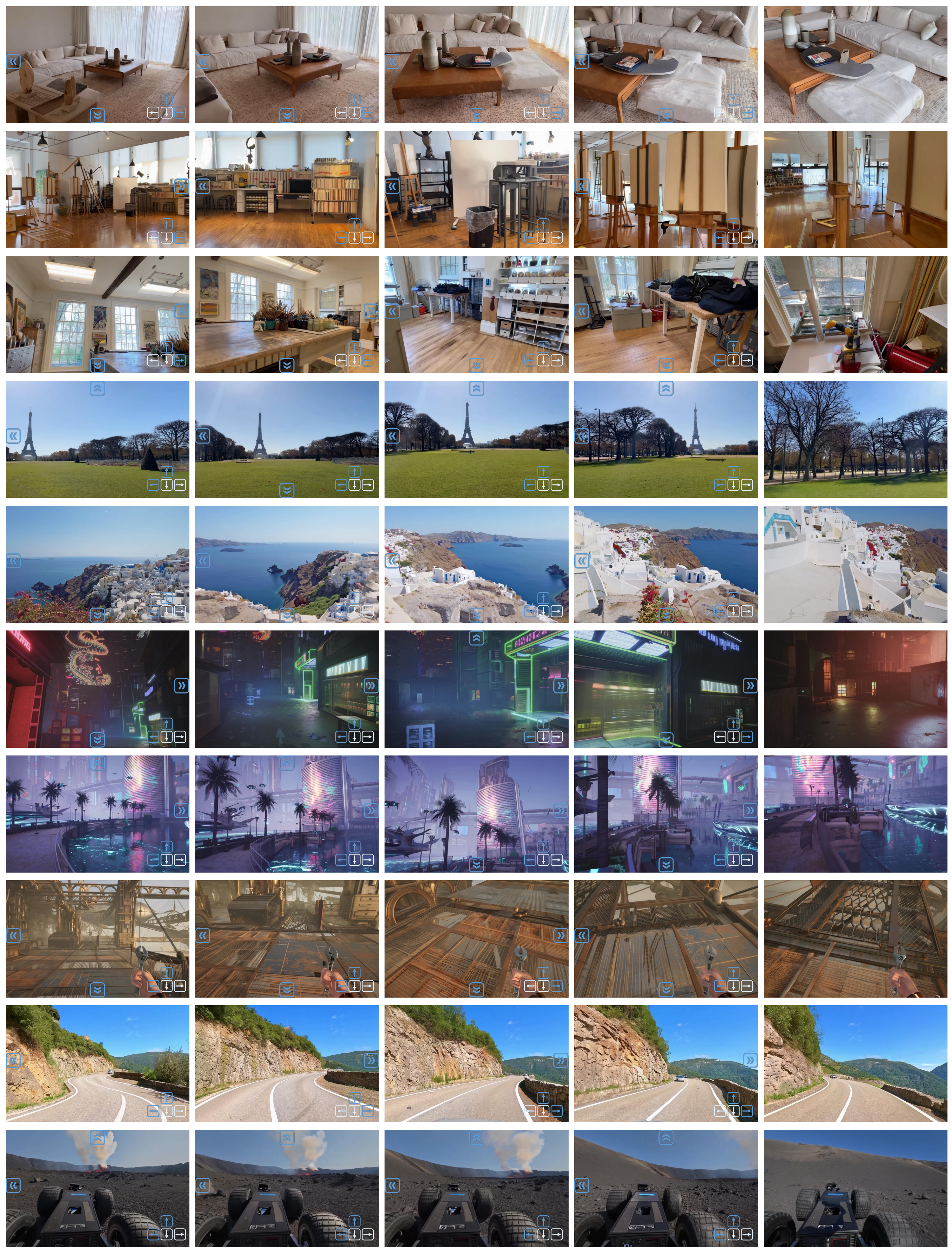}
    }
    \caption{World exploration results of \Ours.
    The overlaid buttons on each frame represent the camera movement or viewpoint change instructions from the current frame to the next.
    }
    \label{fig:example_explore}
\end{figure}

\paragraph{Qualitative Analysis.\!\!\!}
To thoroughly evaluate our model’s exploration capabilities across diverse and unseen scenarios, we focus on the pure-text input under User-Interactive Mode setting, which allows comprehensive assessment of both initial frame generation and subsequent stepwise exploration. 
As shown in Figure~\ref{fig:example_explore}, our \Ours preserves stable scene layouts and natural camera transitions over extended trajectories, bridging the gap between discrete frame generation and continuous environmental rendering. 
Under the User-Interactive Mode, the model effectively follows user instructions step by step, demonstrating precise control and adaptive scene evolution. 
\Ours further maintains strong spatial reasoning and context retention, ensuring coherent scene structure and temporal continuity throughout the exploration trajectory.
Collectively, these results validate the effectiveness of \Ours in achieving long-range consistency, controllable interactivity, and real-virtual scene integration, establishing a principled direction toward unified multimodal world understanding.

\subsection{Embodied Manipulation}

\paragraph{Quantitative Evaluation.}
For quantitative assessment, we sample a subset from the validation set, comprising 5 different robotic embodiments and over 50 distinct tasks ranging from 3 to 13 steps. To evaluate the generalization capability of \Ours, we employ Gemini 2.5 Flash Image to perturb the initial position of the robotic arm and modify environmental conditions—including lighting, object appearance, scene layout, camera viewpoint, and background. Additionally, we capture 10 real-world images from various indoor scenes to further extend the diversity of evaluation scenarios. In total, the resulting benchmark consists of 331 samples (10 real captured, 109 sampled, and 192 synthesized). Similar to the evaluation setup of other interleaved multimodal tasks, we use ChatGPT as an automated judge to compare \Ours against Gemini 2.5 Flash Image.
For each manipulation sample, ChatGPT assesses the following criteria: \textit{Subtask Skill Clarity}, \textit{Text–Image Alignment}, \textit{Task Execution Progress}, \textit{Image Quality}, \textit{Background Consistency}, and \textit{Physical Law Consistency}. As shown in Table~\ref{tab:eval_winrate}, \Ours significantly outperforms Gemini 2.5 Flash Image in both in-domain and out-of-distribution settings. This indicates that \Ours is capable of generating manipulation tasks with coherent interaction processes and high physical plausibility. These results demonstrate that our model not only comprehends interaction mechanics and object affordances, but also excels in modeling scenario dynamics involving rigid and deformable objects, liquids, and complex backgrounds.

\begin{figure}[t]
    \centering
    \makebox[\textwidth][c]{
        \includegraphics[width=1.1\linewidth]{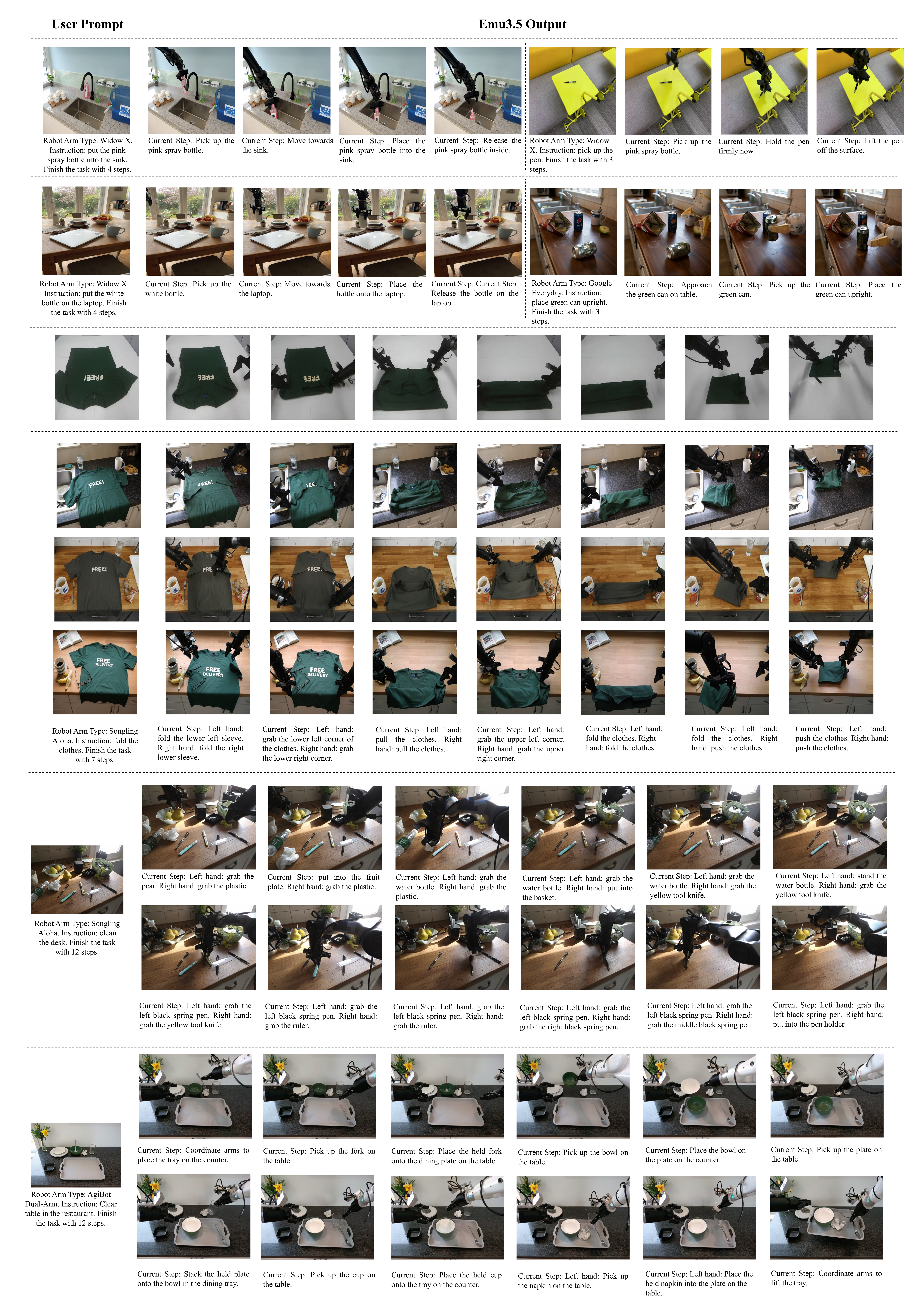}
    }
    \caption{Embodied manipulation results of \Ours.}
    \label{fig:example_vla}
\end{figure}

\paragraph{Qualitative Analysis.}
As shown in Figure~\ref{fig:example_vla}, \Ours demonstrates strong performance and generalization in embodied manipulation generation with an interleaved subtask-keyframe format. Specifically, the framework supports multiple viewpoints, skills, and embodiments, achieving superior performance in background consistency, execution integrity, and physical law adherence.
All initial frames in Figure~\ref{fig:example_vla} are generated by Gemini 2.5 Flash Image (except the first row of cloth folding), featuring unseen backgrounds, objects, and layouts. Notably, the initial positions of the robotic arms are not provided in the prompt images, yet the model can predict reasonable arm placements and subsequent interaction processes.
More importantly, the cloth folding process of the Songling Aloha robot illustrates the model's semantic alignment capability across different layouts. In the first row, the cloth folding scenario matches the training samples, where the collar of the T-shirt faces downward toward the closer side of the table. In evaluation cases, however, the T-shirt is placed upside down with the collar facing the farther table side. Despite this alteration and the significant variance of camera view and scene background, the model successfully recognizes semantic components of the T-shirt (\eg, sleeves, collar, cloth corners) and completes the folding process in strict accordance with the subtask plans.
These results demonstrate the strong generalization capacity and effectiveness of \Ours in long-horizon embodied planning and manipulation. The framework provides a unified world model that accommodates arbitrary skills, scenarios, and embodiments while maintaining strict compliance with physical laws, as well as ensuring integrity and consistency throughout the manipulation process.

\begin{table}

\centering

\small
\setlength{\tabcolsep}{4pt}
\resizebox{\textwidth}{!}{\begin{tabular}{l c c  c c c c c c c c  c c  c c }

\toprule

\multirow{2}{*}{\textbf{Method}} & \multirow{2}{*}{\textbf{Type}} & \multirow{2}{*}{\textbf{Factor}} & \multicolumn{8}{c}{{\bf Text}(\%)} & \multicolumn{2}{c}{\bf Face}  & \multicolumn{2}{c}{\textbf{General}} \\
\cmidrule(lr){4-11} \cmidrule(lr){12-13} \cmidrule(lr){14-15}
& & &{T-ACC$_{s}\uparrow$} & {T-ACC$_{me}\uparrow$} &{T-ACC$_{l}\uparrow$} & {T-ACC$_{m}\uparrow$} & {T-NED$_{s}\uparrow$} & {T-NED$_{me}\uparrow$} & {T-NED$_{l}\uparrow$} &{T-NED$_{m}\uparrow$}& {F-Sim$_{s}\uparrow$} &{F-Sim$_{m}\uparrow$}  & rFID{$\downarrow$} & LPIPS{$\downarrow$} \\

\midrule
\multicolumn{13}{c}{\textit{Resolution: 512 $\times$ 512}} \\
\midrule

VQGAN~\cite{esser2021taming} & VQ &$16\times$ &0.15 & 0.76 & 17.45 &6.12 & 5.20 & 8.99 & 37.77 & 17.32 & 0.08 &0.11 & 1.19 & 0.13 \\ 
Chameleon~\cite{team2024chameleon} & VQ &$16\times$ & 0.60  &2.67 & 31.39 &11.55 & 7.63  & 17.82 & 54.95 &26.80  &0.13  &0.21 &1.03 &0.12 \\
LlamaGen~\cite{llamagen} & VQ &$16\times$ & 0.67 &3.93 & 40.43 &15.01 & 7.76  & 20.17 & 63.39 &30.44  &0.11 &0.17 &0.68 &0.10 \\
VAR~\cite{VQ:VAR} & RQ &$16\times$ & 3.71  &20.59  & 63.62 & 29.32& 18.01 & 49.56 & 82.44 &50.0  &0.14 &\textbf{0.24} &0.60 &0.09 \\
TokenFlow~\cite{VQ:tokenflow}& RQ &$16\times$ &1.06 & 6.27 & 44.88 &17.40  &10.00 &28.39 &68.07  &35.49  &0.11 &0.16  &1.26 &0.10 \\
O-MAGVIT2~\cite{VQ:openmagvit2} & LFQ &$16\times$  & 1.40 & 9.51 & 54.04 &21.65 & 10.79 & 33.15 & 74.94    &39.63  &0.13 &0.22 &0.55 &0.09 \\
O-MAGVIT2(pretrain)~\cite{VQ:openmagvit2} & LFQ &$16\times$  & 3.02 & 16.25 & 62.72 &27.33  & 16.87 &44.50 & 80.48  &47.28   &0.13 &0.22  &0.45 &0.08 \\
\midrule
\Ours-Vanilla Decoder & IBQ & $16\times$ & 12.99 & 39.59& 71.97 & 41.52 & 39.92 & 68.87 & 87.38 & 65.39 & 0.14 & 0.22 & \textbf{0.42} & \textbf{0.08} \\
\Ours-Diffusion Decoder & IBQ & $16\times$ & \textbf{18.61} & \textbf{53.22} & \textbf{81.50} & \textbf{51.11} & \textbf{43.81} & \textbf{75.96} & \textbf{91.78} & \textbf{70.52} & \textbf{0.14} & \underline{0.22} & 0.49 & 0.10\\
\bottomrule
\end{tabular}}
\vspace{5pt}
\caption{Comparison of discrete tokenizers. ``$_{s}$'', ``$_{me}$'', ``$_{l}$'' and ``$_{m}$'' denote the average metrics for small-, medium-, large-scale instances and all scales, respectively. Factor denotes the downsampling ratio.} 
\label{tab:tokenizer_reconstruction}
\vspace{-3mm}
\end{table}

\subsection{Tokenizer Reconstruction}
\begin{figure}[t]
    \centering
    \makebox[\textwidth][c]{%
        \includegraphics[width=1.0\linewidth]{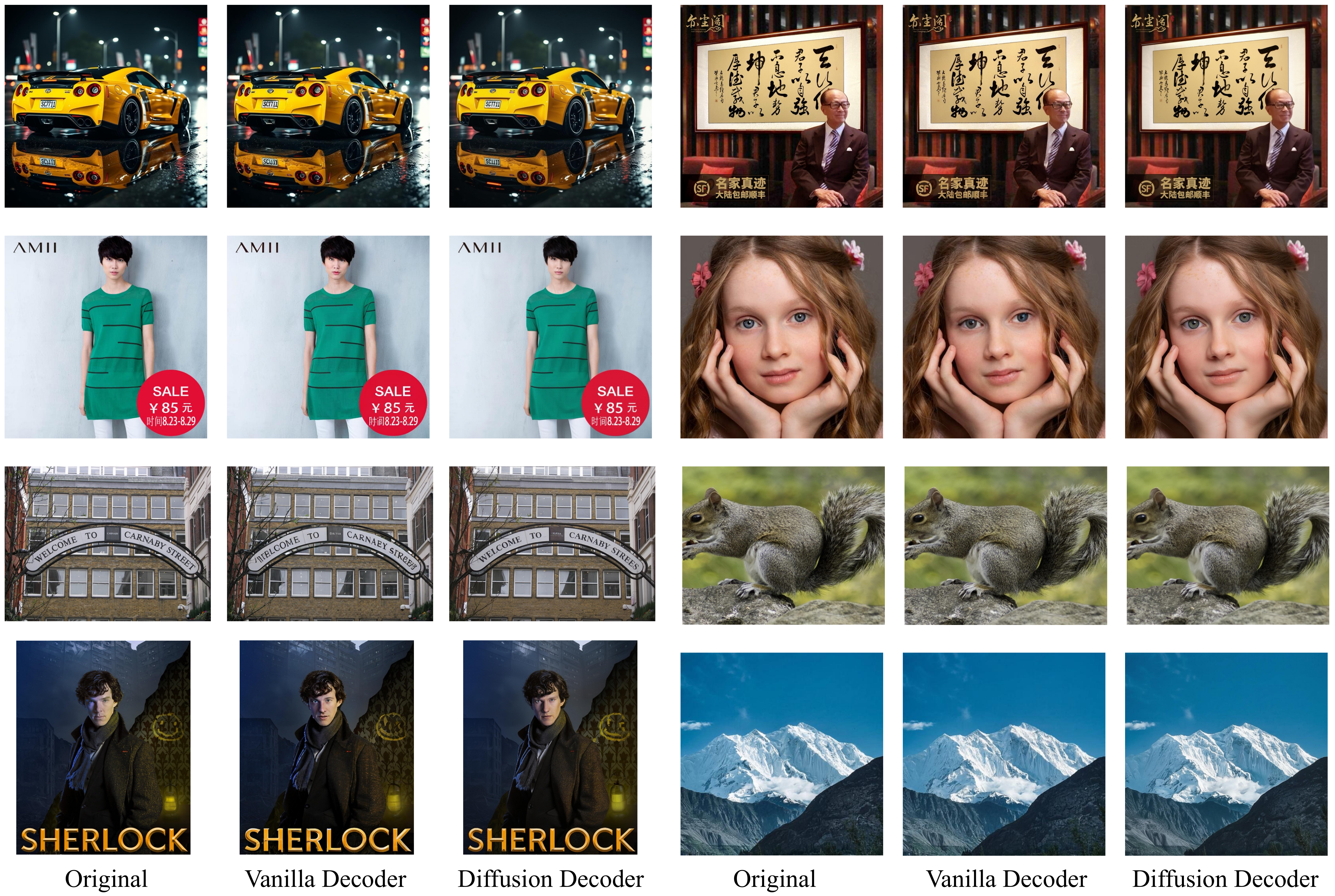}
    }
    \caption{Qualitative results of reconstructions with different decoders, \ie the vanilla image decoder and the diffusion-based one.}
    \label{fig:example_reconstruction}
\end{figure}

To quantify the representational capability of our tokenizer, we perform comprehensive evaluations on various perspectives. Specifically, we adopt Tokbench~\cite{VQ:Tokbench} to measure the textual and facial performance, and simultaneously curate an high quality 60k evaluation set comprising across diverse domains for general reconstruction ability assessment. As shown in Table~\ref{tab:tokenizer_reconstruction}, \Ours demonstrates superior representations on textual and facial features (\eg, 51.11 T-ACC$_{m}$, 70.52 T-NED$_{m}$) as well as competitive performance on general domain reconstructions. Besides, we provide visualization of different decoders in Figure~\ref{fig:example_reconstruction}. 
By employing diffusion-based decoder which generates 2$\times$ resolution for restoration, the visual fidelity and detailed perception get consistent improved. 

\subsection{\Ursa}

\begin{table}[htbp]
\centering
\resizebox{\textwidth}{!}{
\renewcommand{\arraystretch}{1.0}
\setlength{\tabcolsep}{6pt}
\begin{tabular}{l|c|c|c|c|cc|c|c|c}
\toprule
\multirow{2}{*}{\textbf{Model}} & 
\multirow{2}{*}{\textbf{\#Params}} & 
\multirow{2}{*}{\textbf{Resolution}} & 
\multirow{2}{*}{\textbf{\#Gen Tokens}} & 
\multirow{2}{*}{\textbf{Gen Method}} & 
\multicolumn{2}{c|}{\textbf{Inference Time (s)}} & 
\multirow{2}{*}{\textbf{GenEval~\cite{ghosh2024geneval}}} & 
\multirow{2}{*}{\textbf{DPG-Bench~\cite{hu2024ella}}} &
\textbf{GEdit-Bench-EN~\cite{liu2025step1x}} \\
\cmidrule(lr){6-7}
 & & & & & \textbf{Naive} & \textbf{FlagScale\cite{flagscale2025}} & & & G\_O \\
\midrule
Emu3 & 8B & $720\times720$ & 8,100 &AR & 260 & 68 & 0.66 & 80.60 & -- \\
\Ours & 34B & $1024\times1024$ & 4,096 & AR & 512 & 120 & \textbf{0.86} & \textbf{88.26} & \textbf{7.59} \\
\Ours & 34B & $1024\times 1024$ & 4,096 & DiDA& 22 & 10 & \textbf{0.86} & 87.46  & 7.56 \\
\bottomrule
\end{tabular}
}
\vspace{6pt}
\caption{Evaluation of inference speed and performance for \Ours variants and Emu3 on T2I and X2I tasks. We report the inference time for text-to-image task.}
\vspace{-6pt}
\label{tab:model_comparison_t2i}
\end{table}

We compare different inference variants of \Ours to highlight the effectiveness of the \Ursaabbr. As shown in Table~\ref{tab:model_comparison_t2i}, \Ursaabbr achieves up to $20\times$ faster inference while maintaining comparable performance to the AR baseline on both text-to-image tasks (\eg GenEval~\citep{ghosh2024geneval}, DPG-Bench~\citep{hu2024ella}) and image editing tasks (\eg GEdit-Bench~\citep{liu2025step1x}). With the support of FlagScale~\citep{flagscale2025}, \Ursaabbr can generate a 4,096-token image in only 10s, achieving inference speed on par with continuous diffusion counterpart using fast sampling strategies~\citep{ddim, dpmsolver}.
These results underscore the superior inference efficiency of \Ursaabbr without compromising visual quality.

\section{Conclusion, Limitations and Future Work}

In this work, we present \Ours, a large-scale multimodal world model that natively predicts the next state across interleaved vision and language. By training end-to-end with a unified next-token prediction objective on over 10 trillion multimodal tokens and post-training with large-scale multimodal reinforcement learning, \Ours establishes a new foundation for long-horizon vision-language generation and reasoning. The model demonstrates strong native multimodal capabilities, achieving state-of-the-art performance in any-to-image (X2I) generation, text-to-image generation, and complex interleaved tasks. Beyond perception and generation, \Ours exhibits generalizable world-modeling abilities that enable long-horizon prediction, world exploration, and embodied interaction, marking a key step toward general-purpose multimodal intelligence.

We openly release both the model and its full development journey, including the data pipeline, powerful tokenizer, native multimodal pre-training, unified post-training, and the discrete diffusion adaptation method for efficient inference. We hope this comprehensive release will serve as a foundation for future research in large-scale world models.

In future work, we plan to advance \Ours along several directions.

\begin{itemize}
    \item \textbf{Improved Tokenizer:} While \Ours significantly improves the compactness and representation of the tokenizer compared to Emu3, it still requires 1024 tokens to encode a 512$\times$512 image. We aim to further enhance the compression ratio and reconstruction fidelity.
    \item \textbf{Inference Acceleration:} The proposed discrete diffusion adaptation (DiDA) accelerates autoregressive prediction by up to 20$\times$ without sacrificing performance. We plan to continue exploring acceleration methods to further lower inference latency.
    \item \textbf{Comprehensive Evaluation:} The new abilities of \Ours, such as visual narratives and visual guidance, are still under-evaluated. We hope to establish systematic benchmarks of quantitative and human evaluations to better assess long-horizon multimodal generation.
    \item \textbf{Advanced Prompting:} \Ours already demonstrates strong instruction-following ability in any-to-image generation. We plan to explore more multimodal prompting strategies to probe the limits of the world model.
    \item \textbf{Embodied Agents:} With its powerful world modeling capabilities in embodied manipulation and world exploration, we will extend \Ours to serve as the foundation for generalizable embodied agents interacting in the physical world.

\end{itemize}

\section{Authors and Contributions}

* core contributors ordered by Chinese surname stroke count

\subsection*{Tokenizer}
Fan Zhang*, Zhuoyan Luo*, Xu Huang*

\subsection*{Data}
Yueze Wang*, Wenxuan Wang*,  Yufeng Cui*, Chengyuan Wang,  Xinghang Li, Honghao Chen, Fan Zhang, Yang Liu,  Zhuoyan Luo, Jinsheng Wang

\subsection*{Pre-training}
Yufeng Cui*,  Wenxuan Wang, Yueze Wang, Zhuoyan Luo, Yingli Zhao, Fan Zhang, Zecheng Hao, Wenxuan Ma

\subsection*{Downstream Post-training}
Yueze Wang*, Wenxuan Wang*, Chengyuan Wang*,  Jirong Liu*, Yang Liu*, Honghao Chen*, Xinghang Li*, Fan Zhang*, Zhuoyan Luo*, Yufeng Cui*

\subsection*{Unified Post-training}
Jinsheng Wang*, Yang Liu*, Jirong Liu*, Yufeng Cui, Chengyuan Wang, Wenxuan Wang, Honghao Chen

\subsection*{Discrete Diffusion Adaptation}
Haoge Deng*, Fan Zhang*, Ting Pan, Yingli Zhao, Xianduo Li

\subsection*{Infrastructure}
Yingli Zhao*, Xianduo Li, Zhuo Chen, Yulong Ao

\subsection*{Senior Leads}
Zhongyuan Wang, Tiejun Huang

\subsection*{Project Lead}
Xinlong Wang

\section*{Acknowledgements}
\vspace{-1em}
We thank Zhiyu Li, Mengsi Lv, Tengfei Pan,  Quanyue Ma, Liang Sun, Chunlei Men, Shaokai Nie, Xuan Liu, Bochao Zhang,  Guang Liu, Liangdong Wang,  Hua Zhou, Yaohui Chen, Jinxin Xie, Yance Jiao, Yiwen Shao, Xiaodan Liu, Jiawei Ding, Jingshu Zheng, Zhuang Yu, Jiaqi Xu, Huanran Wang, Wenfei Xie, Naiwen Xu for their support for the Emu3.5 project.

\newgeometry{top=0.75cm,bottom=2cm}
\restoregeometry

\newgeometry{top=0.75cm,bottom=2cm}
\restoregeometry

\newgeometry{top=0.75cm,bottom=2cm}
\restoregeometry

\newgeometry{top=0.75cm,bottom=2cm}
\restoregeometry

\newgeometry{top=0.75cm,bottom=2cm}
\restoregeometry

\newgeometry{top=0.75cm,bottom=2cm}
\restoregeometry

\newgeometry{top=0.75cm,bottom=2cm}
\restoregeometry

\newgeometry{top=0.75cm,bottom=2cm}
\restoregeometry

{
\bibliographystyle{plain}
\bibliography{main}
}

\end{document}